\documentclass[journal]{IEEEtran}

\usepackage[T1]{fontenc}
\usepackage{cite}
\usepackage{amsmath,amsfonts,amssymb,amsthm,mathtools,bm}
\usepackage{graphicx}
\usepackage[caption=false,font=footnotesize]{subfig}
\usepackage{booktabs}
\usepackage{tabularx}
\usepackage{array}
\usepackage{siunitx}
\usepackage[ruled,vlined]{algorithm2e}
\usepackage{enumitem}
\usepackage{placeins}
\usepackage{mathrsfs}
\usepackage[hidelinks]{hyperref}

\theoremstyle{plain}
\newtheorem{theorem}{Theorem}[section]

\newtheorem{proposition}[theorem]{Proposition}
\newtheorem{definition}[theorem]{Definition}

\newcommand{\argmin}{\mathop{\rm arg~min}\limits}

\title{Geometric Data Perturbation with Noisy-Anchor Alignment \\for Privacy-Preserving Collaborative Learning}

\author{Keiyu Nosaka, Yamato Suetake, Yuichi Takano, Yukihiko Okada, and Akiko Yoshise%
\thanks{Keiyu Nosaka and Yamato Suetake are with the Graduate School of Science and Technology, University of Tsukuba, Tsukuba, Ibaraki 305-8573, Japan (e-mail: s2430118@u.tsukuba.ac.jp; s2620455@u.tsukuba.ac.jp).}%
\thanks{Yuichi Takano, Yukihiko Okada, and Akiko Yoshise are with the Institute of Systems and Information Engineering, University of Tsukuba, Tsukuba, Ibaraki 305-8573, Japan, and with the Center for Artificial Intelligence Research, Tsukuba Institute for Advanced Research, University of Tsukuba, Tsukuba, Ibaraki 305-8577, Japan (e-mail: ytakano@sk.tsukuba.ac.jp; okayu@sk.tsukuba.ac.jp; yoshise@sk.tsukuba.ac.jp).}%
\thanks{Corresponding author: Akiko Yoshise.}%
\thanks{This work was partially supported by the Japan Society for the Promotion of Science KAKENHI under Grant Numbers JP22K18866, JP23K26327, JP25KJ0701, and JP26K22555.}
\thanks{This work has been submitted to the IEEE for possible publication.
Copyright may be transferred without notice, after which this version
may no longer be accessible.}}

\begin{document}
\maketitle

\begin{abstract}
Geometric data perturbation enables one-shot representation sharing for privacy-preserving collaborative learning: each participant applies a secret distance-preserving transformation to its private data and uploads the resulting representation to a central analyst. We study analyst–participant collusion, in which a colluding participant discloses its data and transformation to help the analyst reconstruct another participant's data. Independent participant-specific transformations block direct inversion through a disclosed common transformation but leave uploads in incompatible coordinate systems, degrading pooled learning. Data Collaboration analysis restores compatibility by aligning transformed copies of a common anchor matrix withheld from the analyst. We show that, when the centered anchor matrix has full column rank, a colluder who discloses it enables exact recovery of every participant's transformation and inversion of noiseless private representations. Adding noise to private-data representations leaves this transformation-recovery channel intact and reduces leakage at a substantial utility cost. Instead, we perturb the anchor representations: each participant perturbs only its transformed anchor representation, preserving the geometry of its private-data upload while turning known-anchor transformation recovery into a noisy estimation problem. The analyst estimates the alignment using a spectral estimator for a generalized orthogonal Procrustes problem. We analyze recovery attacks against this protocol and compare both noise placements on the CelebA and VGGFace2 facial image datasets. Under the evaluated collusion attacks, noisy-anchor alignment retains higher downstream accuracy at low identity-linkage levels. Participant-count experiments examine the utility gains and limitations of larger collaborations at comparable measured linkage.
\end{abstract}


\section{Introduction}
\label{sec:introduction}

Pooling data across organizations can substantially improve predictive performance and robustness, especially when models must generalize across heterogeneous populations. Yet in many domains, raw data cannot simply be shared or centralized due to legal, contractual, operational, and governance constraints \cite{eu2016gdpr,oecd2021easd,govuk2022datasharing,vanpanhuis2014barriers}. Privacy-preserving collaborative learning (PPCL) offers one response to this barrier: it enables organizations to participate in joint model development while constraining what other organizations, or a central analyst, can learn about their records \cite{jiang2021lightweight,shao2024selectivefd}.

We focus on a PPCL regime that we call \emph{one-shot representation sharing (OSRS)}. In this regime, each participant (the data-holding organization) uploads an obfuscated data matrix—a locally computed intermediate representation of its records—to a central analyst, who then trains downstream models using the uploaded representations. This upload-once, train-centrally workflow is useful when multi-round coordination is impractical, for example, because participants have limited compute resources, connectivity is unreliable, or deployment environments impose strict communication constraints \cite{bonawitz2019flscale,sugawara2026dccfl}. Whereas standard iterative federated learning algorithms repeatedly exchange and aggregate model updates, such as parameters or gradients \cite{mcmahan2017fedavg,kairouz2021flproblems}, OSRS communicates the representation itself and requires no subsequent optimization rounds with the participants.

A central design question in OSRS is how to \emph{obfuscate} the uploaded representation so that it remains useful for downstream learning while the risk of reconstruction or re-identification stays limited. Differential privacy is the standard approach when no restrictions are placed on the attacker's auxiliary knowledge. It bounds how much any single protected contribution can influence the released result while still allowing patterns shared across contributions to be learned, typically by adding calibrated random noise to the representation before release~\cite{dwork2006calibrating,dwork2014algorithmic}. The protection comes at a cost in utility: in high-dimensional regimes, the noise required for strong protection substantially distorts the geometric and statistical structure that downstream models exploit, and the privacy--utility trade-off becomes pronounced~\cite{balle2018gaussian,xu2021ppml}.

These utility costs motivate obfuscation mechanisms that preserve more of the structure needed for learning, particularly when restricting auxiliary knowledge is justified. In this study, we focus on collaboration among a small number of identifiable organizations or IoT devices. In such collaborations, participants can specify who receives the data, what uses are permitted, and what information is exchanged, while each participant retains control of its original records, thereby enabling disclosure to be assessed within a defined setting~\cite{nist2023deidentifying}. \emph{Geometric data perturbation (GDP)} is attractive here because its distance-preserving transformations retain useful statistical and geometric structure while remaining computationally lightweight~\cite{liu2006randomprojection,chen2011gdp}.

The privacy limitations of GDP are well studied. Given enough independent input--representation pairs, an attacker can identify the transformation, at least partially~\cite{liu2006attacker,chen2007resilient,giannella2013knowninputs}, and then recover private data from the uploaded representations. In the classical single-data-owner setting, an attacker must obtain such pairs from outside the protocol. In our multi-participant setting, by contrast, the protocol itself can supply the required information: each participant knows its own private data and transformation parameters, and the analyst holds every uploaded representation. If the analyst and one participant pool their information, a shared transformation parameter can be paired with another participant's corresponding representation, opening a direct channel for transformation recovery. The threat is especially relevant to PPCL because the analyst keeps the uploaded representations, so they remain available for retrospective analysis if a participant's private state is disclosed later. This concern leads us to study \emph{analyst--participant collusion}, in which the analyst and one participant combine their views to infer another participant's private data.

The consequences of collusion depend critically on how transformation parameters are shared across participants. We consider two regimes. In \emph{common-transform GDP (C-GDP)}, all participants use the same transform, whereas in \emph{independent-transform GDP (I-GDP)}, each participant uses a separately generated transform \cite{jiang2021lightweight}. Under C-GDP, a colluding participant can disclose the common transform to the analyst, enabling exact inversion of the other participants' representations. Under I-GDP, disclosing one participant's transform does not directly expose the other participants' transforms.

The isolation of I-GDP comes at a practical cost. Although each transformation preserves within-participant geometry, representations produced under different transforms do not share a common coordinate system. Directly pooling them can therefore create incompatible feature-label relationships and degrade conventional downstream learning. This leads to the central design problem addressed in this paper: how can OSRS restore cross-participant comparability without disclosing alignment information that enables exact recovery of participant-specific transforms under analyst-participant collusion?

Data Collaboration (DC) analysis provides a natural candidate for restoring comparability. It introduces a synthetic anchor matrix whose rows serve as common reference records. The same anchor matrix is shared among all participants but is withheld from the analyst. Each participant applies its own transformation to both its private data and the anchor matrix and sends the resulting representations to the analyst. Because the anchor representations correspond to the same underlying reference records, the analyst can use them to map the participant-specific private representations into a common coordinate system~\cite{imakura2020dca,imakura2021accuracy,nosaka2026odc,kawakami2024geneigen}. DC-style alignment, therefore, appears to combine the isolation of I-GDP with conventional pooled learning.

We show, however, that DC-style anchor-aligned GDP (AA-GDP) inherits the same privacy-utility trade-offs as C-GDP under analyst-participant collusion. When the centered anchor matrix has full column rank, alignment produces exactly the C-GDP form with the reference participant's transformation. Moreover, under collusion, the analyst can recover the transformations of the non-colluding participants from their anchor representations, given the stated rank and identifiability conditions. This recovery enables inversion of the corresponding private-data representations.

A natural approach is to add noise to the private-data representations~\cite{yamashiro2024dcdp}. We establish that the equivalence to C-GDP continues to hold when this noise is independent, isotropic, and Gaussian: alignment preserves the Gaussian noise law and produces the corresponding noisy C-GDP form. Because the anchor representations remain noiseless, exact transformation recovery also remains possible under the stated collusion conditions. Together, these results establish that AA-GDP variants inherit the same privacy--utility trade-offs as their corresponding C-GDP variants under this collusion model.

These findings suggest perturbing the alignment signal rather than the private-data signal. We therefore add noise to the anchor representations while leaving the private-data representations unnoised. The resulting noisy multi-participant alignment can be formulated as the generalized orthogonal Procrustes problem~\cite{ling2023gpm}. Anchor noise preserves the within-participant Euclidean geometry of each private-data upload while converting exact, known-anchor transformation recovery into a noisy estimation problem. It creates tunable, attack-dependent operating points along the trade-off between cross-participant alignment accuracy and resistance to the evaluated reconstruction attacks. On the CelebA~\cite{liu2015celeba} and VGGFace2~\cite{cao2018vggface2} facial image datasets, we observe a more favorable measured privacy--utility trade-off than with C-GDP using private-data noise. Overall, anchor noise restores useful cross-participant comparability, retains participant-specific transformations, and replaces exact anchor-based inversion with noise-limited estimation of transformations.

\subsection{Contributions}
\label{subsec:contributions}

\begin{figure*}[!t]
\centering
\includegraphics[width=\linewidth]{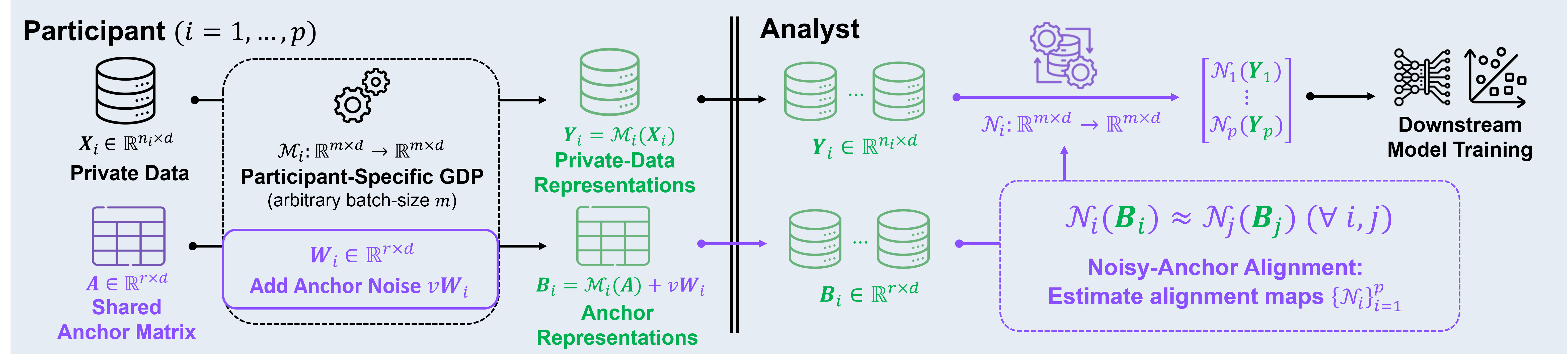}
\caption{Overview of the NAA-GDP pipeline.}
\label{fig:aa_igdp_anchor_noise_overview}
\end{figure*}

Fig.~\ref{fig:aa_igdp_anchor_noise_overview} summarizes our noisy-anchor-aligned GDP (NAA-GDP) pipeline. Each participant applies a participant-specific GDP mechanism to its private-data matrix and the shared anchor matrix, adds noise only to the resulting anchor representation, and uploads both representations. From the noisy anchor representations, the analyst estimates participant-specific alignment maps, maps the private-data representations into a shared coordinate system, and trains a downstream model.

We summarize our contributions as follows.

\begin{enumerate}

    \item \textbf{C-GDP Equivalence of AA-GDP:}
    We establish equivalence between AA-GDP and C-GDP under analyst-participant collusion, including cases where independent isotropic Gaussian noise is added to the private data representations. When the centered anchor matrix has full column rank, and the anchor representations are noiseless, alignment produces exactly the C-GDP form with the reference participant's transformation and preserves the Gaussian noise law. Together with exact transformation recovery under collusion, this establishes that AA-GDP variants inherit the same privacy--utility trade-offs as their corresponding C-GDP variants under the stated collusion model.

    \item \textbf{Alignment with Anchor Noise:} Motivated by the aforementioned equivalence, we introduce noisy-anchor-aligned GDP (NAA-GDP), which perturbs the anchor representations instead of the private-data uploads. We formulate alignment from noisy anchors as a generalized orthogonal Procrustes problem, estimate the alignment using the spectral estimator~\cite{ling2023gpm}, and analyze reconstruction under analyst-participant collusion.

    \item \textbf{Empirical Privacy--Utility Characterization:} Experiments on the CelebA and VGGFace2 facial image datasets compare 100 accepted realizations per noise placement, with downstream classifiers trained afresh and identity linkage evaluated by a frozen auditor. NAA-GDP retains higher mean utility than its private-data noise counterpart in the lower-linkage region under the evaluated attacks. Nested experiments with 2 to 50 participants on both datasets examine how utility changes with collaboration size at comparable measured identity linkage.

\end{enumerate}

\subsection{Organization}
\label{subsec:organization}

Section~\ref{sec:preliminaries} presents the notation, the OSRS protocol, the collusion threat model, and the GDP and anchor-aligned baselines. Section~\ref{sec:limits_existing} shows that existing AA-GDP is equivalent to C-GDP under collusion. Section~\ref{sec:pa_i_gdp} develops NAA-GDP, Section~\ref{sec:empirics} reports the experiments, and Section~\ref{sec:conclusion} concludes; proofs, additional reconstruction attacks, and their experimental comparison are given in the supplementary material.

\section{Preliminaries}
\label{sec:preliminaries}

This section fixes the notation, system model, and threat model, and restates the comparison mechanisms in a common notation. GDP and its additive-noise extension follow the geometric-perturbation literature~\cite{liu2006randomprojection,chen2011gdp}; independent participant-specific transforms follow~\cite{jiang2021lightweight}; generic anchor-based collaboration follows~\cite{imakura2020dca,imakura2021accuracy}; the noiseless anchor-aligned GDP protocol and its reference-participant alignment procedure follow~\cite{nosaka2026odc}; and private-data noise in DC follows~\cite{yamashiro2024dcdp}.

\subsection{Notation}
\label{subsec:notation}
For a positive integer $k$, we write $[k] := \{1,2,\dots,k\}$. We denote by $\mathbb{R}^{p\times q}$ the set of real $p\times q$ matrices. $(\bm{M})_{k,l}$ denotes the row $k$, column $l$ entry of the matrix $\bm{M}$. $\bm I_p \in\mathbb{R}^{p\times p}$ is the identity matrix and $\mathbf{1}_{p}\in\mathbb{R}^{p}$ is the all-ones vector. For a matrix $\bm{M}$, $\bm{M}^\top$ denotes the transpose, $\bm{M}^\dagger$ the Moore--Penrose pseudoinverse, $\|\bm{M}\|_{\mathrm{F}}$ the Frobenius norm, and $\mathrm{tr}(\bm{M})$ the matrix trace. We denote the orthogonal group in $\mathbb{R}^{p}$ by \(\mathcal{O}(p):=\{\bm{O}\in\mathbb{R}^{p\times p}:\bm{O}^\top\bm{O}=\bm{I}_p\}\). For a matrix $\bm{M}$, its (thin) singular value decomposition (SVD) is written as $\bm{M}=\bm{U}\bm{\Sigma}\bm{V}^\top$. We write $\begin{bmatrix} \bm{M}_i \end{bmatrix}_{i\in [k]}$ for $k$ vertically stacked matrices. Equivalently,
\[
\begin{bmatrix}
\bm{M}_i
\end{bmatrix}_{i\in [k]} :=
\begin{bmatrix}
\bm{M}_1 \\
\vdots \\
\bm{M}_k
\end{bmatrix}.
\]
For a matrix $\bm{M}\in\mathbb{R}^{p\times q}$, we write $\bm{m}:=\frac{1}{p}\bm{M}^\top\mathbf{1}_p$ for its column-mean vector and $\overline{\bm{M}}:=\bm{M}-\mathbf{1}_p\bm{m}^\top$ for its centered version.

\subsection{One-Shot Representation Sharing}
\label{subsec:One-Shot}

We study the \emph{one-shot representation sharing (OSRS)} regime in the PPCL setting. Assume a total of $p$ participants and let \(\mathcal{P}\coloneqq[p]\) denote the participant set. We assume a horizontal partition: all participants use the same \(d\)-dimensional feature schema, and their $n_i$-sample record sets are disjoint. For supervised tasks, participant \(i\) also holds a label vector \(\bm{L}_i\in\mathbb{R}^{n_i}\). We suppress these labels from most of the notation and assume that they are analyst-visible and included in the same one-shot message; the reconstruction target studied here is the feature matrix \(\bm{X}_i\in\mathbb{R}^{n_i\times d}\). 

\begin{definition}[One-Shot Representation Sharing]
\label{def:one_shot_representation_sharing}
In the one-shot representation sharing regime, the global record-by-feature data matrix $\bm{X}\in\mathbb{R}^{n\times d}$ is partitioned by records across $p$ participants:
\begin{equation}
\bm{X}
=
\begin{bmatrix}
\bm{X}_i
\end{bmatrix}_{i\in \mathcal{P}}
\qquad
\bm{X}_i\in\mathbb{R}^{n_i\times d},
\qquad
\sum_{i\in\mathcal{P}} n_i = n .
\end{equation}
Participant $i$ holds the private submatrix $\bm{X}_i$. For any batch size $m$, participant $i$ applies a local row-wise obfuscation mechanism
\[
    \mathcal{M}_i:\mathbb{R}^{m\times d}\rightarrow\mathbb{R}^{m\times h},
\]
which maps each record to an $h$-dimensional representation. Applying this mechanism to its local data yields
\begin{equation}
    \bm{Y}_i
    =
    \mathcal{M}_i(\bm{X}_i)
    \in\mathbb{R}^{n_i\times h}.
\end{equation}
Each participant communicates with the analyst exactly once by uploading \(\bm{Y}_i\) and the target vector \(\bm{L}_i\in\mathbb{R}^{n_i}\). The analyst concatenates the uploaded representations as
\begin{equation}
\bm{Y}
=
\begin{bmatrix}
\bm{Y}_i
\end{bmatrix}_{i \in \mathcal{P}}
\in\mathbb{R}^{n\times h},
\end{equation}
and uses $\bm{Y}$ for downstream analysis, such as model training.
\end{definition}

We specifically consider the \emph{analyst-participant collusion setting}, defined as follows.

\begin{definition}[Analyst-Participant Collusion]
\label{def:analyst_participant_collusion}
Consider a one-shot representation-sharing protocol with $p\geq2$ participants and a central analyst. Let $c\in\mathcal{P}$ denote the colluding participant. Under \emph{analyst-participant collusion}, the attacker comprises the central analyst and participant $c$.

We assume that participant $c$ discloses both $\bm{X}_c$ and $\mathcal{M}_c$ in full. The analyst observes all uploaded representations $\{\bm{Y}_i:i\in\mathcal{P}\}$. Hence, the attacker's view is
\begin{equation}
    \left(
    \{\bm{Y}_i : i\in\mathcal{P}\},
    (\bm{X}_c,\mathcal{M}_c)
    \right).
\end{equation}
The attacker also knows the public protocol specification, including the functional forms of the obfuscation mechanisms. The attacker does not directly know the private datasets or the participant-specific obfuscation mechanisms of the non-colluding participants.

The attacker's objective is to recover, or infer sensitive information about, the private dataset $\bm{X}_i\in\mathbb{R}^{n_i\times d}$ of a non-colluding target participant $i\in\mathcal{P}\setminus\{c\}$.
\end{definition}

\subsection{Geometric Data Perturbation}
\label{subsec:GDP}

\emph{Geometric Data Perturbation (GDP)} is a classical approach to privacy-preserving data mining in which a data owner releases a geometrically transformed version of its dataset for analysis~\cite{liu2006randomprojection,chen2011gdp}. We adopt the standard variant that composes an orthogonal transformation with a translation.

\begin{definition}[Geometric Data Perturbation]
\label{def:GDP}
For any batch size $m$, a Geometric Data Perturbation (GDP) mechanism with parameters $(\bm{O},\Psi)$ is the row-wise obfuscation map $\mathcal{M}:\mathbb{R}^{m\times d}\rightarrow\mathbb{R}^{m\times d}$ given by
\begin{equation}
    \mathcal{M}(\bm{S})
    =
    \bm{S}\bm{O}
    +
    \mathbf{1}_{m}\Psi^\top,
    \qquad
    \bm{S}\in\mathbb{R}^{m\times d},
\end{equation}
where $\bm{O}\in\mathcal{O}(d)$ is an orthogonal matrix and $\Psi\in\mathbb{R}^{d}$ is a translation vector. Thus, the same orthogonal transformation and translation are applied to every record.
\end{definition}

Because $\bm{O}$ is orthogonal, the map preserves pairwise Euclidean distances and, hence, the geometric structure on which downstream analyses rely. The translation is not needed for this preservation; instead, it plays a protective role: an orthogonal transformation alone leaves the origin fixed, so records near it are only weakly perturbed, and an unknown translation hides where this region lies. On its own, however, a translation offers no protection~\cite{chen2011gdp}.

Classical GDP assumes a single data owner and therefore does not specify how perturbation parameters $(\bm{O},\Psi)$ should be coordinated among the participants of a PPCL protocol. The most direct baseline is for all participants to use the same parameters; we refer to this regime as \emph{Common-Transform Geometric Data Perturbation}.

\begin{definition}[Common-Transform GDP]
\label{def:common_GDP}
Common-Transform Geometric Data Perturbation (C-GDP) is a one-shot representation-sharing regime in which, for any batch size $m$, all participants use the same GDP mechanism $\mathcal{M}:\mathbb{R}^{m\times d}\rightarrow\mathbb{R}^{m\times d}$ with common parameters $(\bm{O},\Psi)$. Each participant $i\in\mathcal{P}$, holding private data $\bm{X}_i\in\mathbb{R}^{n_i\times d}$, uploads
\begin{equation}
    \bm{Y}_i
    =
    \mathcal{M}(\bm{X}_i)
    =
    \bm{X}_i\bm{O}
    +
    \mathbf{1}_{n_i}\Psi^\top.
\end{equation}
\end{definition}

Under analyst-participant collusion (\textbf{Definition~\ref{def:analyst_participant_collusion}}), sharing the parameters opens an immediate and exact inversion channel: the disclosed mechanism $\mathcal{M}_c=\mathcal{M}$ contains $(\bm{O},\Psi)$, so the attacker recovers the private data of every non-colluding participant exactly as
\[
    \bm{X}_i
    =
    \left(\bm{Y}_i-\mathbf{1}_{n_i}\Psi^\top\right)\bm{O}^\top,
    \qquad i\in\mathcal{P}\setminus\{c\}.
\]

Classical security analyses of GDP consider known-input, known-sample, and Independent Component Analysis (ICA)-based attacks~\cite{chen2011gdp,giannella2013knowninputs}. Of these, the known-input attack, in which the attacker observes a subset of original records together with their perturbed versions, is the closest to analyst-participant collusion. The difference is that the colluder discloses the mechanism itself, not merely input--output pairs from which it would have to be estimated, which is why the inversion above is exact. A simple countermeasure studied in the GDP literature is to superimpose random noise on the geometric perturbation~\cite{chen2011gdp}. In our multi-participant setting, this idea yields a regime in which all participants share the same transformation but add independent, participant-specific noise to their transformed records before uploading them; we refer to this regime as \emph{C-GDP (private-data noise)}.

\begin{definition}[C-GDP with Private-Data Noise]
\label{def:CGDP_PN}
C-GDP (private-data noise) uses the common-transform GDP mechanism $\mathcal{M}$ of \textbf{Definition~\ref{def:common_GDP}} and adds independent noise to each private-data representation: participant $i\in\mathcal{P}$ draws a noise matrix $\bm{W}_i\in\mathbb{R}^{n_i\times d}$ from a chosen base distribution and uploads
\begin{equation}
    \bm{Y}_i
    =
    \mathcal{M}(\bm{X}_i)
    +
    \sigma\bm{W}_i
    =
    \bm{X}_i\bm{O}
    +
    \mathbf{1}_{n_i}\Psi^\top
    +
    \sigma\bm{W}_i,
\end{equation}
where $\sigma\geq0$ is the private-data-noise scale. The matrices $\bm{W}_1,\ldots,\bm{W}_p$ are drawn independently across participants and independently of $(\bm{O},\Psi)$; setting $\sigma=0$ recovers C-GDP.
\end{definition}

Under the same collusion, the inversion above no longer returns the records themselves: for any non-colluding target $i\in\mathcal{P}\setminus\{c\}$,
\[
    (\bm{Y}_i-\mathbf{1}_{n_i}\Psi^\top)\bm{O}^\top
    =
    \bm{X}_i+\sigma\bm{W}_i\bm{O}^\top,
\]
so the additive noise, rather than the secrecy of the common transform, is the only thing protecting the records.

In this study, the entries of each $\bm{W}_i$ are i.i.d.\ $\mathcal{N}(0,1)$, so $\sigma$ is the standard deviation of each added noise entry; when a formal differential-privacy baseline is required, $\sigma$ can be calibrated to the sensitivity of the released representation, the neighboring relation, and the target privacy parameters~\cite{dwork2014algorithmic,balle2018gaussian}. Because the noise protects the uploads only by distorting them, however, its protection comes at a direct cost in utility: in high-dimensional regimes, the noise scale required for strong protection erodes the geometric and statistical structure that downstream models exploit, producing a steep privacy--utility trade-off~\cite{balle2018gaussian,xu2021ppml}.

Because C-GDP is vulnerable to analyst-participant collusion and additive noise protects it only at a steep cost to utility, a natural alternative is to give each participant its own independently sampled geometric perturbation~\cite{jiang2021lightweight}. We refer to this regime as \emph{Independent-Transform Geometric Data Perturbation}.

\begin{definition}[Independent-Transform GDP]
\label{def:Independent_GDP}
Independent-Transform GDP (I-GDP) is a one-shot representation-sharing regime in which, for any batch size $m$, each participant $i\in\mathcal{P}$ uses its own GDP mechanism $\mathcal{M}_i:\mathbb{R}^{m\times d}\rightarrow\mathbb{R}^{m\times d}$ with parameters $(\bm{O}_i,\Psi_i)$, where the pairs $(\bm{O}_1,\Psi_1),\ldots,(\bm{O}_p,\Psi_p)$ are sampled independently across participants. Participant $i$, holding private data $\bm{X}_i\in\mathbb{R}^{n_i\times d}$, uploads
\begin{equation}
    \bm{Y}_i
    =
    \mathcal{M}_i(\bm{X}_i)
    =
    \bm{X}_i\bm{O}_i
    +
    \mathbf{1}_{n_i}\Psi_i^\top.
\end{equation}
\end{definition}

I-GDP removes the shared-parameter attack channel: the parameters of one participant carry no information about those of another, so the colluder's disclosure of $(\bm{O}_c,\Psi_c)$ does not, by itself, enable inversion of any other participant's upload. Its cost, instead, is a loss of cross-participant comparability. Each transformation preserves the Euclidean geometry of its own participant's data up to a rigid transformation, but the uploaded blocks lie in unrelated coordinate systems. Conventional learners, which assume that each feature coordinate has a consistent meaning across samples, may therefore lose accuracy when trained on the concatenated matrix $\bm{Y}$. Expressive neural networks can partially adapt to this mismatch, but they require additional model capacity, and their performance falls as the number of participants grows, even when the amount of available data increases~\cite{jiang2021lightweight}. Restoring this comparability---so that additional participants improve rather than degrade pooled learning---is the purpose of the anchor alignment described next.

\subsection{Data-Collaboration-Style Anchor Alignment}
\label{subsec:dca}

\emph{Data Collaboration (DC)} analysis is an anchor-based extension of OSRS~\cite{sugawara2026dccfl,imakura2020dca,nosaka2026odc,imakura2023nridca}. In DC analysis, the participants share a synthetic anchor matrix (typically randomly generated) among themselves, transform it with the same local map as their private data, and upload both representations, while the anchor matrix itself is withheld from the analyst.

\begin{definition}[DC-Style OSRS]
\label{def:dca_anchor_setting}
A DC-style OSRS is an extension to the one-shot representation-sharing regime in which every participant $i\in\mathcal{P}$, holding its private data $\bm{X}_i\in\mathbb{R}^{n_i\times d}$, has access to a common anchor matrix $\bm{A}\in\mathbb{R}^{r\times d}$. The anchor matrix is shared on the participant side: it is not revealed to the analyst in the honest protocol. For any batch size $m$, participant $i$ has a local obfuscation map
\[
    \mathcal{M}_i:\mathbb{R}^{m\times d}\rightarrow\mathbb{R}^{m\times h},
\]
which it applies to both its private-data matrix and the shared anchor matrix, uploading
\[
    \bm{Y}_i=\mathcal{M}_i(\bm{X}_i),
    \qquad
    \bm{B}_i=\mathcal{M}_i(\bm{A}).
\]
\end{definition}

Because participant-specific transformations place the uploaded representations in different coordinate systems, DC-style anchor alignment uses the uploaded anchor representations as a common reference for reconciling them. From the collection $\{\bm B_i\}_{i\in\mathcal P}$, the analyst constructs, for each participant $i$ and any batch size $m$, a row-wise alignment map
\[
    \mathcal N_i:\mathbb R^{m\times h}\to\mathbb R^{m\times h}
\]
that carries representations from participant $i$'s coordinate system into a shared collaboration space, applies it to the corresponding private-data representation,
\[
    \bm Z_i = \mathcal N_i(\bm Y_i),
\]
and pools the aligned representations for downstream analysis. Although $\mathcal N_i$ is estimated from the anchor representations alone, applying it to $\bm Y_i$ is legitimate because participant $i$ generates $\bm B_i$ and $\bm Y_i$ with the same local map $\mathcal M_i$. We call the GDP-specific instantiation considered below, which combines independent GDP transformations with the alignment procedure~\cite{nosaka2026odc}, \emph{Anchor-Aligned GDP (AA-GDP)}.

\begin{definition}[Anchor-Aligned GDP]
\label{def:aaigdp}
Anchor-Aligned GDP (AA-GDP) is the DC-style OSRS of \textbf{Definition~\ref{def:dca_anchor_setting}} in which, for any batch size $m$, every participant's local map is its I-GDP mechanism of \textbf{Definition~\ref{def:Independent_GDP}},
\[
    \mathcal{M}_i(\bm{S})
    =
    \bm{S}\bm{O}_i
    +
    \mathbf{1}_{m}\Psi_i^\top,
    \qquad
    \bm{S}\in\mathbb{R}^{m\times d},
\]
with $(\bm{O}_i,\Psi_i)\in\mathcal{O}(d)\times\mathbb{R}^{d}$ sampled independently across participants; in particular, $h=d$. The uploads thus become
\begin{align}
    \bm{Y}_i
    &=
    \mathcal{M}_i(\bm{X}_i)
    =
    \bm{X}_i\bm{O}_i
    +
    \mathbf{1}_{n_i}\Psi_i^\top,
    \label{eq:aaigdp_private_upload}
    \\
    \bm{B}_i
    &=
    \mathcal{M}_i(\bm{A})
    =
    \bm{A}\bm{O}_i
    +
    \mathbf{1}_{r}\Psi_i^\top .
    \label{eq:aaigdp_anchor_upload}
\end{align}
\end{definition}

Because the rows of every $\bm{B}_i$ are the images of the same anchor records in the same order, the analyst can estimate the relative transformation between any two participants' coordinate systems from the anchor representations alone, without access to $\bm{A}$ itself. 

\textbf{Algorithm~\ref{alg:aaigdp}} specializes the generic DC workflow to AA-GDP accordingly. For each participant $i$, let $\bm{b}_i:=\frac{1}{r}\bm{B}_i^\top\mathbf{1}_r$ denote the column-mean vector of $\bm{B}_i$, and let $\overline{\bm{B}}_i:=\bm{B}_i-\mathbf{1}_r\bm{b}_i^\top$ denote its centered version. A reference participant $s\in\mathcal{P}$ fixes the common coordinate system, and for each $i\in\mathcal{P}$ the analyst rotates the centered anchor representation $\overline{\bm{B}}_i$ onto $\overline{\bm{B}}_s$ by solving the orthogonal Procrustes problem
\[
    \min_{\bm{R}\in\mathcal{O}(d)}
    \left\|\overline{\bm{B}}_i\bm{R}-\overline{\bm{B}}_s\right\|_{\mathrm{F}},
\]
whose solution is $\bm{R}_i=\bm{U}_i\bm{V}_i^\top$, where $\overline{\bm{B}}_i^\top\overline{\bm{B}}_s=\bm{U}_i\bm{\Sigma}_i\bm{V}_i^\top$ is an SVD~\cite{tenberge1977procrustes,nosaka2026odc}. The resulting map $\mathcal{N}_i$ subtracts participant $i$'s anchor-representation mean $\bm{b}_i$, applies the estimated rotation $\bm{R}_i$, and adds the reference participant's mean $\bm{b}_s$; applying it to the private-data representation gives $\bm{Z}_i=\mathcal{N}_i(\bm{Y}_i)$, and the aligned blocks are stacked into $\bm{Z}=\begin{bmatrix}\bm{Z}_i\end{bmatrix}_{i\in\mathcal{P}}$.

\begin{algorithm}[!t]
\footnotesize
\caption{AA-GDP Alignment Procedure}
\label{alg:aaigdp}
\DontPrintSemicolon

\KwIn{Private data $\bm{X}_i\in\mathbb{R}^{n_i\times d}$ for each $i\in\mathcal{P}$; shared anchor matrix $\bm{A}\in\mathbb{R}^{r\times d}$; reference participant $s\in\mathcal{P}$.}
\KwOut{Aligned collaboration matrix $\bm{Z}\in\mathbb{R}^{n\times d}$.}

\BlankLine
\tcc{Participant-Side Obfuscation}
\ForEach{$i\in\mathcal{P}$}{
    Privately sample GDP parameters $\bm{O}_i\in\mathcal{O}(d)$ and $\Psi_i\in\mathbb{R}^{d}$\;
    $\displaystyle \bm{Y}_i = \bm{X}_i\bm{O}_i + \mathbf{1}_{n_i}\Psi_i^\top$\;
    $\displaystyle \bm{B}_i = \bm{A}\bm{O}_i + \mathbf{1}_{r}\Psi_i^\top$\;
    Upload the pair $(\bm{Y}_i,\bm{B}_i)$ to the analyst\;
}

\BlankLine
\tcc{Analyst-Side Alignment}
\ForEach{$i\in\mathcal{P}$}{
    $\displaystyle \bm{b}_i = \frac{1}{r}\bm{B}_i^\top\mathbf{1}_{r}$\;
    $\displaystyle \overline{\bm{B}}_i = \bm{B}_i - \mathbf{1}_{r}\bm{b}_{i}^{\top}$\;
}

\ForEach{$i\in\mathcal{P}$}{
    Compute the SVD $\displaystyle \overline{\bm{B}}_i^\top\overline{\bm{B}}_s = \bm{U}_i\bm{\Sigma}_i\bm{V}_i^\top$\;
    $\displaystyle \bm{R}_i = \bm{U}_i\bm{V}_i^\top$\;
    Define $\displaystyle \mathcal{N}_i(\bm{T})=\left(\bm{T}-\mathbf{1}_{m}\bm{b}_{i}^{\top}\right)\bm{R}_i+\mathbf{1}_{m}\bm{b}_{s}^{\top}$ for $\bm{T}\in\mathbb{R}^{m\times d}$\;
    $\displaystyle \bm{Z}_i = \mathcal{N}_i(\bm{Y}_i)$\;
}

\Return{$\bm{Z} = \begin{bmatrix}
\bm{Z}_i
\end{bmatrix}_{i\in \mathcal{P}}$}
\end{algorithm}

Prior DC work adds calibrated noise to the private-data intermediate representations while keeping the anchor representations noiseless for alignment~\cite{yamashiro2024dcdp}. We instantiate this placement with isotropic Gaussian noise and call the resulting comparison baseline \emph{AA-GDP (private-data noise)}.

\begin{definition}[AA-GDP with Private-Data Noise]
\label{def:AA_I_GDP_DP}
AA-GDP (private-data noise) uses the AA-GDP mechanism $\mathcal{M}_i$ of \textbf{Definition~\ref{def:aaigdp}} for both the private-data matrix and the shared anchor matrix but adds noise only to the private-data representation: participant $i\in\mathcal{P}$ draws a noise matrix $\bm{W}_i\in\mathbb{R}^{n_i\times d}$ with i.i.d.\ $\mathcal{N}(0,1)$ entries and uploads
\begin{align}
    \bm{Y}_i
    &=
    \mathcal{M}_i(\bm{X}_i)
    +
    \sigma\bm{W}_i
    =
    \bm{X}_i\bm{O}_i
    +
    \mathbf{1}_{n_i}\Psi_i^\top
    +
    \sigma\bm{W}_i,
    \\
    \bm{B}_i
    &=
    \mathcal{M}_i(\bm{A})
    =
    \bm{A}\bm{O}_i
    +
    \mathbf{1}_{r}\Psi_i^\top,
\end{align}
where $\sigma\geq0$ is the private-data-noise scale. The noise matrices $\bm W_1,\ldots,\bm W_p$ are drawn independently of one another and independently of all GDP parameters.
\end{definition}

Because the anchor representations $\{\bm{B}_i\}_{i\in\mathcal{P}}$ are unchanged, \textbf{Algorithm~\ref{alg:aaigdp}} returns exactly the same alignment maps as for AA-GDP; the private-data noise enters only through $\bm{Z}_i=\mathcal{N}_i(\bm{Y}_i)$, when each map is applied to the noisy private-data representation.

Finally, for anchor-based protocols, the collusion view of \textbf{Definition~\ref{def:analyst_participant_collusion}} must also specify what happens to the participant-side shared anchor matrix; we use the following extension throughout.

\begin{definition}[Analyst-Participant Collusion of DC]
\label{def:analyst_participant_collusion_DC}
The analyst-participant collusion setting of \textbf{Definition~\ref{def:analyst_participant_collusion}} carries over to a DC-style OSRS with two additions: the analyst observes the uploaded anchor representations $\{\bm{B}_i:i\in\mathcal{P}\}$ alongside $\{\bm{Y}_i:i\in\mathcal{P}\}$, and the colluding participant $c\in\mathcal{P}$ discloses the shared anchor matrix $\bm{A}$ together with $\bm{X}_c$ and $\mathcal{M}_c$. The attacker's view is therefore
\begin{equation}
\label{eq:dcaview}
    \left(
    \{(\bm{Y}_i,\bm{B}_i):i\in\mathcal{P}\},
    (\bm{X}_c,\mathcal{M}_c),
    \bm{A}
    \right),
\end{equation}
while its knowledge of the public protocol specification and its objective against a non-colluding target $i\in\mathcal{P}\setminus\{c\}$ are unchanged.
\end{definition}

The question is how the existing AA-GDP variants fare under this threat model. They regain comparability precisely by reintroducing a participant-side shared object, the anchor matrix $\bm A$, and in the view~\eqref{eq:dcaview} the colluder discloses exactly that object. Section~\ref{sec:limits_existing} shows that this disclosure reopens the channel that I-GDP had closed: alignment returns exactly a C-GDP upload, and the disclosed anchor recovers every non-colluding participant's transformation, so each AA-GDP variant inherits both the utility and the collusion exposure of its C-GDP counterpart.

\section{Equivalence of Existing AA-GDP to C-GDP}
\label{sec:limits_existing}

To establish that existing AA-GDP constructions inherit the vulnerabilities and privacy--utility trade-offs of C-GDP under analyst-participant collusion, we first express AA-GDP in C-GDP form: after alignment, each participant's output is a C-GDP upload with the reference participant's parameters.

\begin{proposition}[C-GDP Form of AA-GDP]
\label{prop:aaigdp_cgdp_form}
Let $\sigma\geq0$, let every participant $i\in\mathcal{P}$ upload $(\bm{Y}_i,\bm{B}_i)$ as in \textbf{Definition~\ref{def:AA_I_GDP_DP}}, and fix a reference participant $s\in\mathcal{P}$. Assume that $\operatorname{rank}(\overline{\bm{A}})=d$ (which requires $r\geq d+1$). Then the analyst-side alignment of \textbf{Algorithm~\ref{alg:aaigdp}}, applied to these uploads, returns, for every $i\in\mathcal{P}$,
\[
\begin{aligned}
    \bm{R}_i
    &=
    \bm{O}_i^\top\bm{O}_s,
    \\
    \bm{Z}_i
    &=
    \bm{X}_i\bm{O}_s
    +
    \mathbf{1}_{n_i}\Psi_s^\top
    +
    \sigma\bm{W}_i\bm{O}_i^\top\bm{O}_s .
\end{aligned}
\]
Consequently, $\bm{Z}=\begin{bmatrix}\bm{Z}_i\end{bmatrix}_{i\in\mathcal{P}}$ is exactly the upload matrix of C-GDP (private-data noise) in \textbf{Definition~\ref{def:CGDP_PN}} with common parameters $(\bm{O}_s,\Psi_s)$ and noise blocks
\[
    \widetilde{\bm{W}}_i
    :=
    \bm{W}_i\bm{O}_i^\top\bm{O}_s,
    \qquad i\in\mathcal{P}.
\]
Moreover, the entries of these blocks are jointly i.i.d.\ $\mathcal{N}(0,1)$ and independent of all GDP parameters.
\end{proposition}

Because $\bm{Z}$ is itself a C-GDP upload matrix, every downstream procedure the analyst runs on it returns the same result as it would under C-GDP, so, on the utility side, AA-GDP offers nothing that C-GDP does not. Privacy is a separate matter, since AA-GDP shares no common parameter pair for a colluder to disclose: each participant draws its own $(\bm{O}_i,\Psi_i)$ and never reveals them. One might therefore hope that AA-GDP retains C-GDP's utility without its exposure, but the following theorem rules this out: once the colluder discloses $\bm{A}$, every uploaded $\bm{B}_i$ becomes a known input--output pair of participant $i$'s transformation, and that is enough to recover it.

\begin{theorem}[Known-Anchor Recovery under Collusion]
\label{thm:known_anchor_recovery}
Consider the analyst-participant collusion setting of \textbf{Definition~\ref{def:analyst_participant_collusion_DC}}, let every participant $i\in\mathcal{P}$ upload $(\bm{Y}_i,\bm{B}_i)$ as in \textbf{Definition~\ref{def:AA_I_GDP_DP}} with $\sigma\geq0$, and assume $\operatorname{rank}(\overline{\bm{A}})=d$. Then, for every non-colluding participant $i\in\mathcal{P}\setminus\{c\}$, the attacker recovers the transformation parameters $(\bm{O}_i,\Psi_i)$ exactly from $(\bm{A},\bm{B}_i)$ and inverts the upload as
\begin{equation}
    \left(
        \bm{Y}_i
        -
        \mathbf{1}_{n_i}\Psi_i^\top
    \right)
    \bm{O}_i^\top
    =
    \bm{X}_i
    +
    \sigma\bm{W}_i\bm{O}_i^\top.
\label{eq:known_anchor_attack}
\end{equation}
\end{theorem}

At $\sigma=0$, the right-hand side of Eq.~\eqref{eq:known_anchor_attack} is $\bm{X}_i$ itself, so the records of every non-colluding participant are recovered exactly; for $\sigma>0$, the additive noise alone, rather than the secrecy of any transformation, stands between the attacker and the records, exactly as in C-GDP (private-data noise). Existing AA-GDP constructions therefore inherit C-GDP's utility (\textbf{Proposition~\ref{prop:aaigdp_cgdp_form}}) together with its exposure under collusion (\textbf{Theorem~\ref{thm:known_anchor_recovery}}), and hence its privacy--utility trade-off: anchor alignment restores comparability across participants, but not the protection that participant-specific transformations were meant to provide, and raising $\sigma$ buys privacy only by degrading the very representations that carry the utility. Both the alignment and the attack run through the anchor representations, and it is there that Section~\ref{sec:pa_i_gdp} intervenes.

\section{Noisy-Anchor Alignment}
\label{sec:pa_i_gdp}

Section~\ref{sec:limits_existing} isolates the anchor representations as the single point at which alignment and reconstruction risk interact: under the full-column-rank condition, they alone determine the alignment maps; together with the disclosed raw anchor matrix, they identify every participant-specific transformation; and private-data noise changes neither property. We therefore add noise to the anchor representations.

\subsection{Protocol and Anchor-Matrix Setup}

\begin{definition}[Noisy-Anchor-Aligned GDP]
\label{def:pa_i_gdp}
Noisy-Anchor-Aligned GDP (NAA-GDP) uses the AA-GDP mechanism $\mathcal{M}_i$ of \textbf{Definition~\ref{def:aaigdp}} for both the private-data matrix and the shared anchor matrix but adds noise only to the anchor representation: participant $i\in\mathcal{P}$ draws a noise matrix $\bm{W}_i\in\mathbb{R}^{r\times d}$ with i.i.d.\ $\mathcal{N}(0,1)$ entries and uploads
\begin{align}
    \bm{Y}_i
    &=
    \mathcal{M}_i(\bm{X}_i)
    =
    \bm{X}_i\bm{O}_i
    +
    \mathbf{1}_{n_i}\Psi_i^\top,
    \\
    \bm{B}_i
    &=
    \mathcal{M}_i(\bm{A})
    +
    v\bm{W}_i
    =
    \bm{A}\bm{O}_i
    +
    \mathbf{1}_r\Psi_i^\top
    +
    v\bm{W}_i,
\end{align}
where $v\geq0$ is the anchor-noise scale. The matrices $\bm W_1,\ldots,\bm W_p$ are sampled independently of one another and independently of the anchor matrix and all GDP parameters.
\end{definition}

We additionally prescribe the shared anchor matrix itself: for $r\geq d+1$, the participants generate, before any release, a single anchor matrix $\bm A\in\mathbb R^{r\times d}$ satisfying
\begin{equation}
    \mathbf{1}_r^\top\bm{A}=\mathbf{0}_d^\top,
    \qquad
    \bm{A}^\top\bm{A}=\bm{I}_d.
    \label{eq:paigdp_centered_orthonormal_anchor}
\end{equation}
Thus, the anchor matrix is centered and has orthonormal columns. The first constraint fixes its centroid at the origin, and the second fixes its magnitude relative to the anchor-noise scale $v$; together they ensure $\overline{\bm A}=\bm A$ and $\operatorname{rank}(\overline{\bm A})=d$, so the full-column-rank condition of Section~\ref{sec:limits_existing} holds by construction. One way to generate such a matrix is to center an i.i.d.\ standard Gaussian matrix and take the orthonormal factor of its sign-corrected thin QR factorization; the experimental construction is given in Section~\ref{subsubsec:datasets}. The same realization of $\bm A$ is distributed to every participant and withheld from the analyst during honest execution.

The honest analyst view therefore consists of $\{\bm{Y}_i,\bm{B}_i:i\in\mathcal{P}\}$ and excludes the anchor matrix $\bm{A}$. Each uploaded anchor representation $\bm{B}_i$ is generated from the same anchor matrix and by the same mechanism $\mathcal{M}_i$ as $\bm{Y}_i$, but is perturbed by $v\bm{W}_i$: the anchor representations still relate every participant's coordinate system to a common reference, now only up to noise, and estimating the alignment from them is the subject of the next subsection.

\subsection{Generalized Orthogonal Procrustes Alignment}
\label{subsec:pa_i_gdp_alignment_maps}

The analyst holds the uploaded anchor representations of \textbf{Definition~\ref{def:pa_i_gdp}} but neither the anchor matrix nor any of the GDP parameters that generated them. What it can exploit is their shared structure: each \(\bm{B}_i\) is a noisy rigid transformation of the same withheld anchor matrix. The alignment is formulated as a generalized orthogonal Procrustes problem that fits a common template \(\bm{C}\in\mathbb{R}^{r\times d}\), which stands in for the anchor matrix, together with participant-specific rotations \(\bm{R}_i\in\mathcal{O}(d)\) and translations \(\boldsymbol{\tau}_i\in\mathbb{R}^{d}\):
\begin{equation}
\label{prob:gopp_full}
\begin{aligned}
\min_{\substack{
    \bm{C},\,\bm{R}_1,\ldots,\bm{R}_p,\\
    \boldsymbol{\tau}_1,\ldots,\boldsymbol{\tau}_p}}
&\quad
\sum_{i=1}^{p}
\left\|
        \bm{C}
        -
        \left(
            \bm{B}_i
            -
            \mathbf{1}_{r}\boldsymbol{\tau}_i^\top
        \right)
        \bm{R}_i
    \right\|_{\mathrm{F}}^2  \\
\text{s.t.}
&\quad
\bm{R}_i\in\mathcal{O}(d),
\quad
i\in\mathcal{P}.
\end{aligned}
\end{equation}
Problem~\eqref{prob:gopp_full}, however, never has a unique solution: the objective cannot distinguish feasible points that differ by a common rigid transformation of the collaboration coordinate system.

\begin{proposition}[Rigid-Motion Invariance]
\label{prop:gopp_invariance}
Let \((\bm{C},\bm{R}_1,\ldots,\bm{R}_p,\boldsymbol{\tau}_1,\ldots,\boldsymbol{\tau}_p)\) be feasible for Problem~\eqref{prob:gopp_full}. For any \(\boldsymbol{s}\in\mathbb{R}^{d}\) and \(\bm{Q}\in\mathcal{O}(d)\), define
\[
\begin{aligned}
    \widetilde{\bm{C}}
    &=
    \left(\bm{C}+\mathbf{1}_{r}\boldsymbol{s}^\top\right)\bm{Q},
    \\
    \widetilde{\bm{R}}_i
    &=
    \bm{R}_i\bm{Q},
    \qquad
    \widetilde{\boldsymbol{\tau}}_i
    =
    \boldsymbol{\tau}_i
    -
    \bm{R}_i\boldsymbol{s},
    \qquad
    i\in\mathcal{P}.
\end{aligned}
\]
Then \((\widetilde{\bm{C}},\widetilde{\bm{R}}_1,\ldots,\widetilde{\bm{R}}_p,\widetilde{\boldsymbol{\tau}}_1,\ldots,\widetilde{\boldsymbol{\tau}}_p)\) is feasible and attains the same objective value.
\end{proposition}

The reparameterization in \textbf{Proposition~\ref{prop:gopp_invariance}} applies the same rigid transformation to every aligned representation, thereby preserving their pairwise Euclidean geometry. A downstream model is trained in the resulting collaboration coordinates. The solution of Problem~\eqref{prob:gopp_full} alone does not identify the secret transformation parameters \((\bm{O}_i,\Psi_i)\), because the common coordinate system is undetermined. This is not a privacy guarantee: such ambiguity can be resolved by additional information. Section~\ref{subsec:paigdp_collusion} analyzes collusion that discloses the shared anchor matrix.

\textbf{Proposition~\ref{prop:gopp_invariance}} also shows that the translation freedom can be removed at no cost: the reparameterization with \(\bm{Q}=\bm{I}_d\) and \(\boldsymbol{s}=-\frac{1}{r}\bm{C}^\top\mathbf{1}_r\) preserves feasibility and the objective value while centering the template,
\[
    \mathbf{1}_r^\top\widetilde{\bm{C}}
    =
    \mathbf{1}_r^\top
    \left(
        \bm{C}
        +
        \mathbf{1}_r\boldsymbol{s}^\top
    \right)
    =
    \mathbf{1}_r^\top\bm{C}
    -
    \mathbf{1}_r^\top\bm{C}
    =
    \mathbf{0}_d^\top.
\]

Under this constraint, the translations and the template can be eliminated in closed form, and Problem~\eqref{prob:gopp_full} reduces to a max-trace problem over the rotations alone. For each participant \(i\in\mathcal{P}\), define the uploaded-anchor mean and the centered anchor representation
\begin{equation}
    \bm{b}_i
    :=
    \frac{1}{r}\bm{B}_i^\top\mathbf{1}_r,
    \qquad
    \overline{\bm{B}}_i
    :=
    \bm{B}_i
    -
    \mathbf{1}_r\bm{b}_i^\top .
    \label{eq:paigdp_anchor_mean}
\end{equation}

\begin{proposition}[Max-Trace Reduction]
\label{prop:gopp_reduction}
Consider Problem~\eqref{prob:gopp_full} with the additional constraint \(\mathbf{1}_r^\top\bm{C}=\mathbf{0}_d^\top\). Every minimizer \((\bm{C}^\star,\bm{R}_1^\star,\ldots,\bm{R}_p^\star,\boldsymbol{\tau}_1^\star,\ldots,\boldsymbol{\tau}_p^\star)\) of the constrained problem satisfies
\[
    \boldsymbol{\tau}_i^\star
    =
    \bm{b}_i,
    \quad
    i\in\mathcal{P},
    \qquad
    \bm{C}^\star
    =
    \frac{1}{p}
    \sum_{i=1}^{p}
    \overline{\bm{B}}_i\bm{R}_i^\star,
\]
and \((\bm{R}_1^\star,\ldots,\bm{R}_p^\star)\) solves
\begin{equation}
    \max_{\bm{R}_1,\ldots,\bm{R}_p\in\mathcal{O}(d)}
    \operatorname{tr}
    \left(
        \bm{R}^\top\bm{G}\bm{R}
    \right),
    \label{prob:gopp_trace_form}
\end{equation}
where \(\bm{R}:=\begin{bmatrix}\bm{R}_i\end{bmatrix}_{i\in[p]}\in\mathbb{R}^{pd\times d}\) stacks the rotations and \(\bm{G}\in\mathbb{R}^{pd\times pd}\) is the block matrix with \((i,j)\) block \(\overline{\bm{B}}_i^\top\overline{\bm{B}}_j\). Conversely, every maximizer of Problem~\eqref{prob:gopp_trace_form}, together with \(\boldsymbol{\tau}_i^\star\) and \(\bm{C}^\star\) as above, minimizes the constrained problem.
\end{proposition}

Problem~\eqref{prob:gopp_trace_form} is exactly the max-trace generalized orthogonal Procrustes problem studied by Ling~\cite{ling2023gpm}; we therefore use the spectral estimator developed there. Let $\bm{H}=\begin{bmatrix}\bm{H}_i\end{bmatrix}_{i\in\mathcal{P}}\in\mathbb{R}^{pd\times d}$ contain orthonormal eigenvectors corresponding to the $d$ largest eigenvalues of $\bm{G}$. The spectral estimator $\widehat{\bm{R}}_i$ is
\begin{equation}
\label{eq:spectralestimator}
    \widehat{\bm{R}}_i
    =
    \bm{U}_i\bm{V}_i^\top,
\end{equation}
where $\bm{H}_i=\bm{U}_i\bm{\Sigma}_i\bm{V}_i^\top$ is the singular value decomposition. 

Using the rotations \(\widehat{\bm{R}}_i\) defined by~\eqref{eq:spectralestimator} and the optimal translations \(\boldsymbol{\tau}_i^\star=\bm{b}_i\) of \textbf{Proposition~\ref{prop:gopp_reduction}}, the analyst defines, for any batch size \(m\),
\begin{equation}
    \mathcal{N}_i(\bm{T})
    =
    \left(
        \bm{T}
        -
        \mathbf{1}_{m}\bm{b}_{i}^{\top}
    \right)
    \widehat{\bm{R}}_i,
    \qquad
    \bm{T}\in\mathbb{R}^{m\times d}.
    \label{eq:paigdp_alignment_map}
\end{equation}
The aligned collaboration matrix is
\begin{equation}
    \bm{Z}
    =
    \begin{bmatrix}
        \mathcal{N}_i(\bm{Y}_i)
    \end{bmatrix}_{i\in\mathcal{P}}
    \in\mathbb{R}^{n\times d}.
    \label{eq:paigdp_collaboration_matrix}
\end{equation}
\textbf{Algorithm~\ref{alg:paigdp_pipeline}} summarizes the complete honest-execution pipeline of NAA-GDP.
\begin{algorithm}[!t]
\footnotesize
\caption{Noisy-Anchor-Aligned GDP}
\label{alg:paigdp_pipeline}
\DontPrintSemicolon
\KwIn{Private data $\bm X_i\in\mathbb R^{n_i\times d}$ for each $i\in\mathcal P$; anchor-noise scale $v\geq0$.}
\KwOut{Aligned collaboration matrix $\bm Z\in\mathbb R^{n\times d}$.}
\BlankLine
\tcc{Participant-Side Obfuscation}
Generate and distribute $\bm A$ satisfying \eqref{eq:paigdp_centered_orthonormal_anchor}\;
\ForEach{$i\in\mathcal P$}{
    Privately sample GDP parameters $\bm O_i\in\mathcal O(d)$ and $\Psi_i\in\mathbb R^d$\;
    Sample $\bm W_i\in\mathbb R^{r\times d}$ with i.i.d.\ $\mathcal N(0,1)$ entries\;
    $\displaystyle \bm Y_i=\mathcal M_i(\bm X_i)=\bm X_i\bm O_i+\mathbf 1_{n_i}\Psi_i^\top$\;
    $\displaystyle \bm B_i=\mathcal M_i(\bm A)+v\bm W_i=\bm A\bm O_i+\mathbf 1_r\Psi_i^\top+v\bm W_i$\;
    Upload $(\bm Y_i,\bm B_i)$ to the analyst\;
}
\BlankLine
\tcc{Analyst-Side Alignment}
\ForEach{$i\in\mathcal P$}{
    $\displaystyle \bm b_i=\frac{1}{r}\bm B_i^\top\mathbf 1_r$ and $\displaystyle \overline{\bm B}_i=\bm B_i-\mathbf 1_r\bm b_i^\top$\;
}
Compute the spectral estimator $\widehat{\bm R}$ in Eq.~\eqref{eq:spectralestimator}\;
\ForEach{$i\in\mathcal P$}{
    Define $\mathcal N_i$ by Eq.~\eqref{eq:paigdp_alignment_map}\;
    $\displaystyle \bm Z_i=\mathcal N_i(\bm Y_i)$\;
}
\Return{$\bm Z=\begin{bmatrix}\bm Z_i\end{bmatrix}_{i\in\mathcal P}$}
\end{algorithm}

\subsection{Reconstruction Risk under Analyst-Participant Collusion}
\label{subsec:paigdp_collusion}

We now depart from honest execution and analyze NAA-GDP under analyst-participant collusion (\textbf{Definition~\ref{def:analyst_participant_collusion_DC}}). For a non-colluding target $i\in\mathcal{P}\setminus\{c\}$, substituting the upload $\bm{B}_i$ of \textbf{Definition~\ref{def:pa_i_gdp}} into \eqref{eq:paigdp_anchor_mean} and using $\mathbf{1}_r^\top\bm{A}=\mathbf{0}_d^\top$ from \eqref{eq:paigdp_centered_orthonormal_anchor} yields
\begin{equation}
    \bm{b}_i
    =
    \frac{1}{r}
    \left(
        \bm{A}\bm{O}_i+\mathbf{1}_r\Psi_i^\top+v\bm{W}_i
    \right)^\top
    \mathbf{1}_r
    =
    \Psi_i+\boldsymbol{\eta}_i,
    \label{eq:pdp_collusion_noisy_mean}
\end{equation}
where $\boldsymbol{\eta}_i:=({v}/{r})\bm{W}_i^\top\mathbf{1}_r$ is the anchor-noise mean. Let $\bm{H}_r:=\bm{I}_r-({1}/{r})\mathbf{1}_r\mathbf{1}_r^\top$ denote the centering matrix, so that $\overline{\bm{B}}_i=\bm{H}_r\bm{B}_i$ by \eqref{eq:paigdp_anchor_mean}. Centering the same upload likewise yields
\begin{equation}
    \overline{\bm{B}}_i
    =
    \bm{H}_r
    \left(
        \bm{A}\bm{O}_i+\mathbf{1}_r\Psi_i^\top+v\bm{W}_i
    \right)
    =
    \bm{A}\bm{O}_i+v\overline{\bm{W}}_i,
    \label{eq:pdp_collusion_centered_noisy_relation}
\end{equation}
where $\overline{\bm{W}}_i:=\bm{H}_r\bm{W}_i$ is the centered anchor noise. 

The attacker knows $(\bm{Y}_i,\bm{B}_i,\bm{A})$ and uses the relations \eqref{eq:pdp_collusion_noisy_mean} and \eqref{eq:pdp_collusion_centered_noisy_relation} to estimate the transformation parameters $(\bm{O}_i,\Psi_i)$ and, ultimately, the private data $\bm{X}_i$.

It is natural to use $\widehat{\Psi}_i=\bm{b}_i$ as the estimate of the translation. For the orthogonal transformation $\bm{O}_i$, we consider the estimator $\widehat{\bm{O}}_i=\bm{U}_i\bm{V}_i^\top$, where $\bm{U}_i\bm{\Sigma}_i\bm{V}_i^\top$ is the SVD of $\bm{A}^\top\overline{\bm{B}}_i$. This estimator is a solution of the orthogonal Procrustes problem~\cite{schonemann1966procrustes}
\begin{equation}
    \widehat{\bm{O}}_{i}
    \in
    \argmin_{\bm{O}\in\mathcal{O}(d)}
    \left\|
        \overline{\bm{B}}_i
        -
        \bm{A}\bm{O}
    \right\|_{\mathrm{F}}^2.
    \label{eq:pdp_op_attack_objective}
\end{equation}
The attacker then reconstructs $\widehat{\bm{X}}_i=\left(\bm{Y}_i-\mathbf{1}_{n_i}\widehat{\Psi}_i^\top\right)\widehat{\bm{O}}_i^\top$.

We refer to $\widehat{\bm{O}}_i$ as the \emph{orthogonal-Procrustes (OP)} estimator. Section~II of the supplementary material defines and compares two additional estimators. The results in Section~II-C support OP as the primary evaluated attack reported here. The evaluated estimators are feasible under the stated threat model but not exhaustive; the leakage they achieve is therefore a lower bound, not a privacy guarantee.

\section{Experimental Evaluation}
\label{sec:empirics}

This section addresses two questions: whether moving the noise from the private-data representations $\bm{Y}_i$ to the anchor representations $\bm{B}_i$ improves the privacy--utility trade-off measured under the evaluated attacks (Section~\ref{subsec:tradeoff}), and whether the utility benefit of alignment still grows with the number of participants once the anchor representations are noisy (Section~\ref{subsec:participant_sensitivity}).

We evaluate these questions using the CelebA dataset~\cite{liu2015celeba} and the VGGFace2 dataset~\cite{cao2018vggface2}. Both datasets provide facial images and identity labels. The \texttt{Smiling} annotations used for the VGGFace2 dataset are supplied separately by MAAD-Face~\cite{terhorst2021maadface}. These datasets were chosen because of their accessibility, scale, and identity labels, useful for privacy evaluation. They also allow direct visual inspection of reconstructed records and the recognizable identity information they retain. 

\subsection{Experimental Setup}
\label{subsec:setup}

All experiments are implemented in Python and run on Google Colab with NVIDIA Tesla T4 GPU; the code is available in the project repository\footnote{https://github.com/KU-Nosaka/noisy-anchor-alignment}.

\subsubsection{Dataset Preprocessing}
\label{subsubsec:datasets}

For the CelebA dataset, each RGB image is center-cropped to a square and resized to $64\times64$ pixels using bilinear interpolation. For the VGGFace2 dataset, the supplied face bounding box is expanded by a factor of $1.3$ in width and height about its center, clipped to the image boundaries, and center-cropped to a square before the same resizing operation. For both datasets, pixel values are scaled to $[0,1]$ and flattened in channel-first order into a $d=12{,}288$-dimensional row. For each dataset separately, we construct a fixed linear feature mapping $f:\mathbb{R}^{d}\rightarrow\mathbb{R}^{400}$, given by $f(\bm x)=\bm x\bm F$, where $\bm F\in\mathbb{R}^{d\times400}$ is an orthonormal basis for the leading right singular subspace obtained by uncentered truncated SVD of 10,000 public images. The public-image identities are disjoint from those represented in all participant training, validation, and test sets. The mapping is shared by all participants, known to the colluding analyst, and held fixed across participant counts and experimental conditions. Applying it row-wise yields $\widetilde{\bm X}_i=\bm X_i\bm F$. Participants apply GDP to these embedded representations. For utility evaluation, the resulting protected or aligned 400-dimensional rows are reshaped into single-channel $20\times20$ arrays, and the downstream CNN classifiers are trained from scratch for each experimental condition.

All GDP and anchor-alignment operations take place in the embedded space. The preceding GDP and alignment results are interpreted for these representations, under their stated assumptions, using the substitution
\[
    (\bm X_i,\bm X,d)
    \leftarrow
    (\widetilde{\bm X}_i,\widetilde{\bm X},h),
\]
where $\widetilde{\bm X}$ stacks the participants' embedded data matrices.

For each dataset, we fix an allocation of 50 participants, each containing 100 distinct facial identities. Within each participant, 70 identities are assigned to training, 15 to validation, and 15 to testing. Reference images reserved for identity-linkage evaluation are excluded from participant uploads and the utility datasets. All retained non-reference images of an identity belong to the same participant and utility split, ensuring that identities are disjoint across participants and across training, validation, and test sets. The privacy--utility comparison in Section~\ref{subsec:tradeoff} uses the first ten participants. The participant-count experiment in Section~\ref{subsec:participant_sensitivity} uses nested subsets comprising the first $p$ participants for $p\in\{2,5,10,20,50\}$.

For each dataset, we use one participant-shared anchor matrix $\bm A\in\mathbb{R}^{r\times h}$ with $r=1600$ and $h=400$, constructed directly in the embedded space independently of the participant data. Specifically, we sample $\bm\Omega\in\mathbb{R}^{r\times h}$ with i.i.d.\ $\mathcal N(0,1)$ entries and define the centering matrix
\[
    \bm H_r=\bm I_r-\frac{1}{r}\mathbf 1_r\mathbf 1_r^\top.
\]
We then compute the sign-corrected thin QR factorization
\[
    \bm H_r\bm\Omega=\bm A\bm R_A,
\]
where $\bm R_A\in\mathbb{R}^{h\times h}$ is upper triangular with positive diagonal. Since $r-1\geq h$, the centered Gaussian matrix has full column rank almost surely. Consequently, $\bm A$ has orthonormal columns orthogonal to $\mathbf 1_r$, satisfying \eqref{eq:paigdp_centered_orthonormal_anchor}. Every participant receives the same $\bm A$; neither $\bm A$ nor $\bm\Omega$ is included in the analyst's honest protocol transcript. Participant-specific rotations are sampled independently from the Haar measure on $\mathcal O(h)$, and translation entries are sampled independently from $\mathcal N(0,1)$. C-GDP uses the reference participant's rotation and translation. Within each dataset, the feature mapping, anchor matrix, participant transformations, data allocation, colluder, and protected participant remain fixed across noise realizations and collaboration sizes.

\subsubsection{Compared Methods}
\label{subsubsec:methods}

Table~\ref{tab:compared_methods} summarizes the four collaborative methods. C-GDP applies a common rotation and translation to all participants' embedded data, whereas I-GDP uses participant-specific transformations without alignment. C-GDP (private-data noise) adds Gaussian noise to the private-data uploads under the common transformation. NAA-GDP instead adds Gaussian noise to the anchor uploads and uses the spectral estimator to align the participants' private-data representations. The noise sweeps compare these two noise placements, with C-GDP and I-GDP without added noise serving as controls. The Local baseline trains a separate CNN for each participant using only that participant's unperturbed embedded data, without pooling, GDP transformation, or alignment.

\begin{table}[!t]
\centering
\caption{Compared methods and their uploads}
\label{tab:compared_methods}
\scriptsize
\renewcommand{\arraystretch}{1.2}
\setlength{\tabcolsep}{1.5pt}
\begin{tabular}{@{}lcc@{}}
\toprule
Method & Private-data upload \(\bm Y_i\) & Anchor upload \(\bm B_i\) \\
\midrule
Local & --- & --- \\
C-GDP & \(\widetilde{\bm X}_i\bm O_s+\mathbf 1_{n_i}\Psi_s^\top\) & --- \\
C-GDP (private-data noise) & \(\widetilde{\bm X}_i\bm O_s+\mathbf 1_{n_i}\Psi_s^\top+\sigma\bm W_i\) & --- \\
I-GDP & \(\widetilde{\bm X}_i\bm O_i+\mathbf 1_{n_i}\Psi_i^\top\) & --- \\
NAA-GDP & \(\widetilde{\bm X}_i\bm O_i+\mathbf 1_{n_i}\Psi_i^\top\) & \(\bm A\bm O_i+\mathbf 1_r\Psi_i^\top+v\bm W_i\) \\
\bottomrule
\end{tabular}
\end{table}

For both the CelebA and VGGFace2 datasets, we vary the Gaussian noise standard deviation within $0<v\leq0.08$ for anchor noise and $0<\sigma\leq3$ for private-data noise. Noise scales are calibrated using identity linkage, with 20 accepted realizations in each linkage interval for each noise placement and collaboration size.

\subsubsection{Utility Evaluation}
\label{subsubsec:utility}

For both datasets, the utility task is binary prediction of the \texttt{Smiling} attribute. The CelebA dataset uses its provided attribute annotations, whereas the VGGFace2 dataset uses the released \texttt{Smiling} annotations from MAAD-Face~\cite{terhorst2021maadface}; images with unknown smiling labels are excluded. Utility is evaluated on images from held-out identities using balanced accuracy, defined as the mean of sensitivity and specificity.

Collaborative utility is evaluated under honest execution. The analyst trains a classifier on the collaboration matrix $\bm Z$. For C-GDP, C-GDP with private-data noise, and I-GDP, this matrix is formed by stacking the participants' private-data uploads. For NAA-GDP, the uploads are first aligned using the noisy anchor representations and spectral estimator for alignment, then stacked. Local classifiers use only their respective participants' embedded data for training, validation, and testing. For each collaboration size $p$, the Local reference is a single balanced accuracy computed from the pooled test confusion counts of the $p$ local models.

\begin{table}[!tbp]
\centering
\caption{CNN architecture for the CelebA and VGGFace2 datasets.}
\label{tab:supp_utility_classifiers}
\footnotesize
\renewcommand{\arraystretch}{1.10}
\setlength{\tabcolsep}{6pt}
\begin{tabular}{@{}lc@{}}
\hline
\textbf{Stage} & \textbf{Output dimensions}\\
\hline
Reshape & $1\times20\times20$\\
Convolution 1, ReLU & $10\times16\times16$\\
Max pooling 1 & $10\times8\times8$\\
Convolution 2, ReLU & $20\times4\times4$\\
Max pooling 2 & $20\times2\times2$\\
Flatten & $80$\\
Fully connected, ReLU & $50$\\
Output logit & $1$\\
\hline
\end{tabular}
\end{table}

Both datasets use the same lightweight CNN architecture summarized in Table~\ref{tab:supp_utility_classifiers}. For each collaboration size $p$, a fresh CNN is trained for each collaborative condition and each participant's Local model, while the public SVD feature mappings remain fixed. Both datasets use the same training settings: 15 epochs, unweighted binary cross-entropy with logits, and Adam with learning rate $10^{-3}$, batch size 256, and no weight decay. We select the checkpoint with the highest validation balanced accuracy, retaining the earliest epoch in a tie. Test predictions are obtained by applying a sigmoid to the output logit and thresholding the resulting probability at $0.5$.

\subsubsection{Privacy Evaluation}
\label{subsubsec:privacy}

Privacy is evaluated under the analyst-participant collusion of \textbf{Definition~\ref{def:analyst_participant_collusion_DC}}. The attacker observes the collaborative uploads in Table~\ref{tab:compared_methods}, knows the public feature mapping $f$, and learns the colluder's embedded records, GDP parameters, and, for NAA-GDP, the shared anchor matrix $\bm A$. For C-GDP and C-GDP (private-data noise), the attacker uses the disclosed common parameters $(\bm O_s,\Psi_s)$ to invert the transformation. For NAA-GDP, the attacker estimates the protected participant's transformation using the OP estimator of Section~\ref{subsec:paigdp_collusion}. I-GDP and Local are included as utility references.

Each attack produces an estimate $\widehat{\widetilde{\bm X}}_i$ of the protected participant's embedded records. For both the CelebA and VGGFace2 datasets, we measure identity leakage through cross-image linkage: matching a reconstructed query image to a different reference image of the same person. Reference images are reserved before the experiments and excluded from participant uploads and the utility datasets.

The estimated SVD representations are decoded using $\bm F^\top$, reshaped into $64\times64$ RGB images, and clipped to $[0,1]$. Reference images undergo the same projection, decoding, and clipping. Both are then resized to $160\times160$ pixels using bilinear interpolation and standardized as $(255x-127.5)/128$. A separate frozen FaceNet encoder, implemented as InceptionResnetV1~\cite{szegedy2017inception} pretrained on CASIA-WebFace~\cite{yi2014learning}, maps these images to $\ell_2$-normalized 512-dimensional representations. Matching uses cosine similarity. The auditor is not fine-tuned on either dataset.

For each dataset and collaboration size $p$, we evaluate one query image from each of the protected participant's 100 identities, covering its training, validation, and test identity splits. Each query is evaluated against ten ten-way galleries. Each gallery contains the identity's reserved reference image and references from nine distinct other identities sampled from the active collaboration. The attacker selects the reference with the greatest cosine similarity. Identity-linkage accuracy is the fraction of correct selections across the resulting 1,000 lineups, with a random-guessing accuracy of $10\%$. The protected identities, query images, and genuine reference images remain fixed across collaboration sizes. Distractor galleries are constructed separately for each $p$ from the active reference pool and remain fixed across methods and noise realizations at that $p$.

\subsection{Privacy--Utility Trade-off}
\label{subsec:tradeoff}

\begin{figure*}[!t]
\centering
\includegraphics[width=\textwidth]{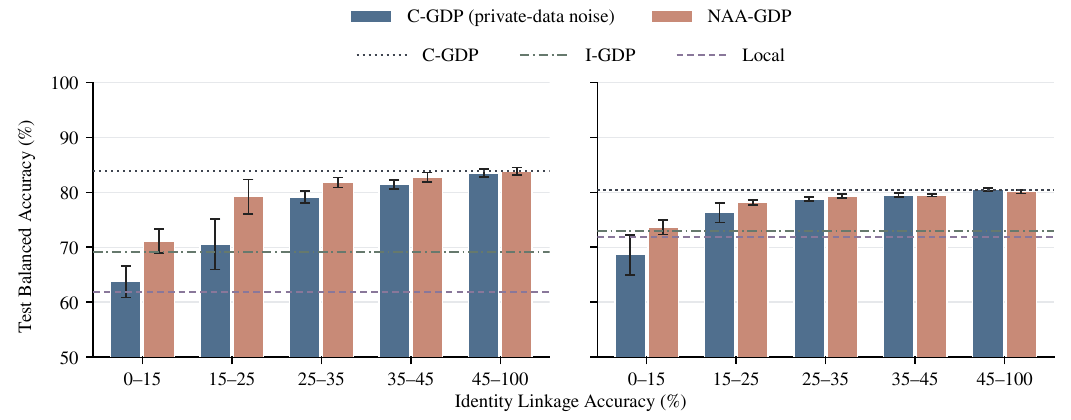}
\caption{Privacy--utility trade-off on the CelebA (left) and VGGFace2 (right) datasets. Bars show mean test balanced accuracy; error bars show one sample standard deviation.}
\label{fig:privacy_utility}
\end{figure*}

Fig.~\ref{fig:privacy_utility} compares the two noise placements at $p=10$ on each dataset. Utility is test \texttt{Smiling} balanced accuracy, with higher values preferred; leakage is cross-image identity-linkage, with lower values preferred. We retain 20 accepted realizations in each of the linkage intervals $[0,15)$, $[15,25)$, $[25,35)$, $[35,45)$, and $[45,100]\%$. Bars show the mean utility within each interval, and error bars show one sample standard deviation across the realizations in that interval. Horizontal lines show C-GDP, I-GDP, and Local utility; the Local value is computed from pooled test predictions of the ten participant-specific models. The utility axis starts at the random guessing level $50\%$.

For the CelebA dataset, Fig.~\ref{fig:privacy_utility} (left) shows higher mean utility for NAA-GDP in all five linkage bins, with the largest differences in the two lowest bins. The noiseless C-GDP control gives $55.7\%$ identity linkage and $83.95\%$ balanced accuracy; I-GDP and Local attain $69.06\%$ and $61.89\%$ balanced accuracy, respectively. In the $[0,15)\%$ linkage bin, NAA-GDP achieves $71.15\%$ mean balanced accuracy, compared with $63.75\%$ for private-data noise, a difference of $7.40$ percentage points. In the $[15,25)\%$ bin, the means are $79.22\%$ and $70.56\%$, a difference of $8.65$ percentage points. The difference decreases to $2.67$ and $1.33$ percentage points in the next two bins, and to $0.29$ percentage points in the $[45,100]\%$ bin, where the means are $83.82\%$ and $83.52\%$.

For the VGGFace2 dataset, Fig.~\ref{fig:privacy_utility} (right) also shows an advantage for anchor noise in the lower-linkage region, but a smaller difference at higher linkage. The noiseless C-GDP control gives $60.2\%$ identity linkage and $80.49\%$ balanced accuracy; I-GDP and Local attain $72.90\%$ and $71.89\%$, respectively. In the $[0,15)\%$ linkage bin, NAA-GDP achieves $73.65\%$ mean balanced accuracy, compared with $68.62\%$ for private-data noise, a difference of $5.03$ percentage points. In the $[15,25)\%$ bin, the means are $78.14\%$ and $76.28\%$, a difference of $1.86$ percentage points. The advantage decreases to $0.55$ percentage points in the $[25,35)\%$ bin. The methods have nearly equal mean utility in the $[35,45)\%$ bin ($79.48\%$ and $79.49\%$), while private-data noise has a $0.44$-percentage-point advantage in the highest bin ($80.54\%$ versus $80.10\%$).

\begin{figure*}[!t]
\centering
\subfloat[CelebA.\label{fig:celeba_reconstructions_panel}]{%
\includegraphics[width=\textwidth]{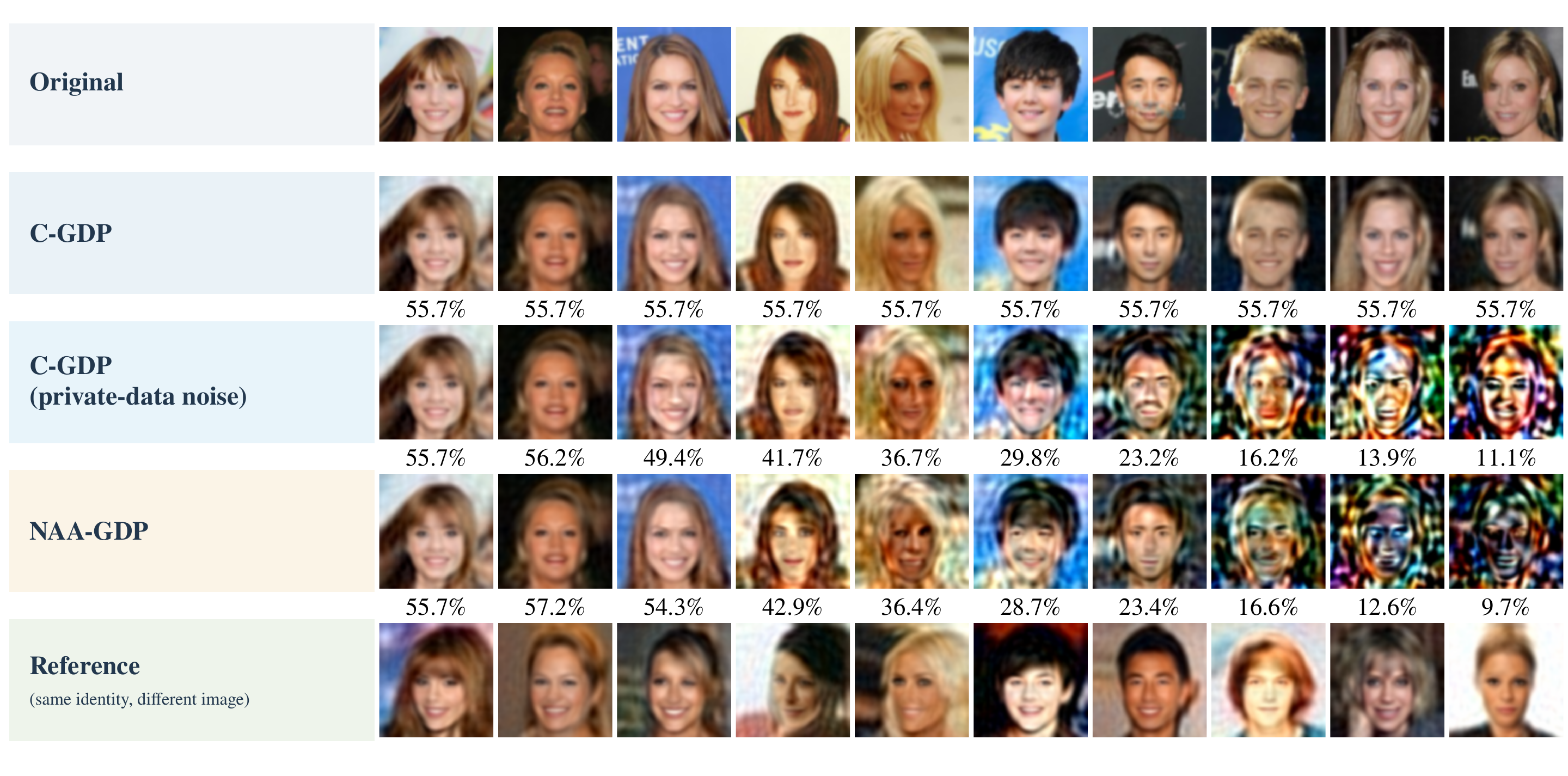}}\\[1mm]
\subfloat[VGGFace2.\label{fig:vggface2_reconstructions_panel}]{%
\includegraphics[width=\textwidth]{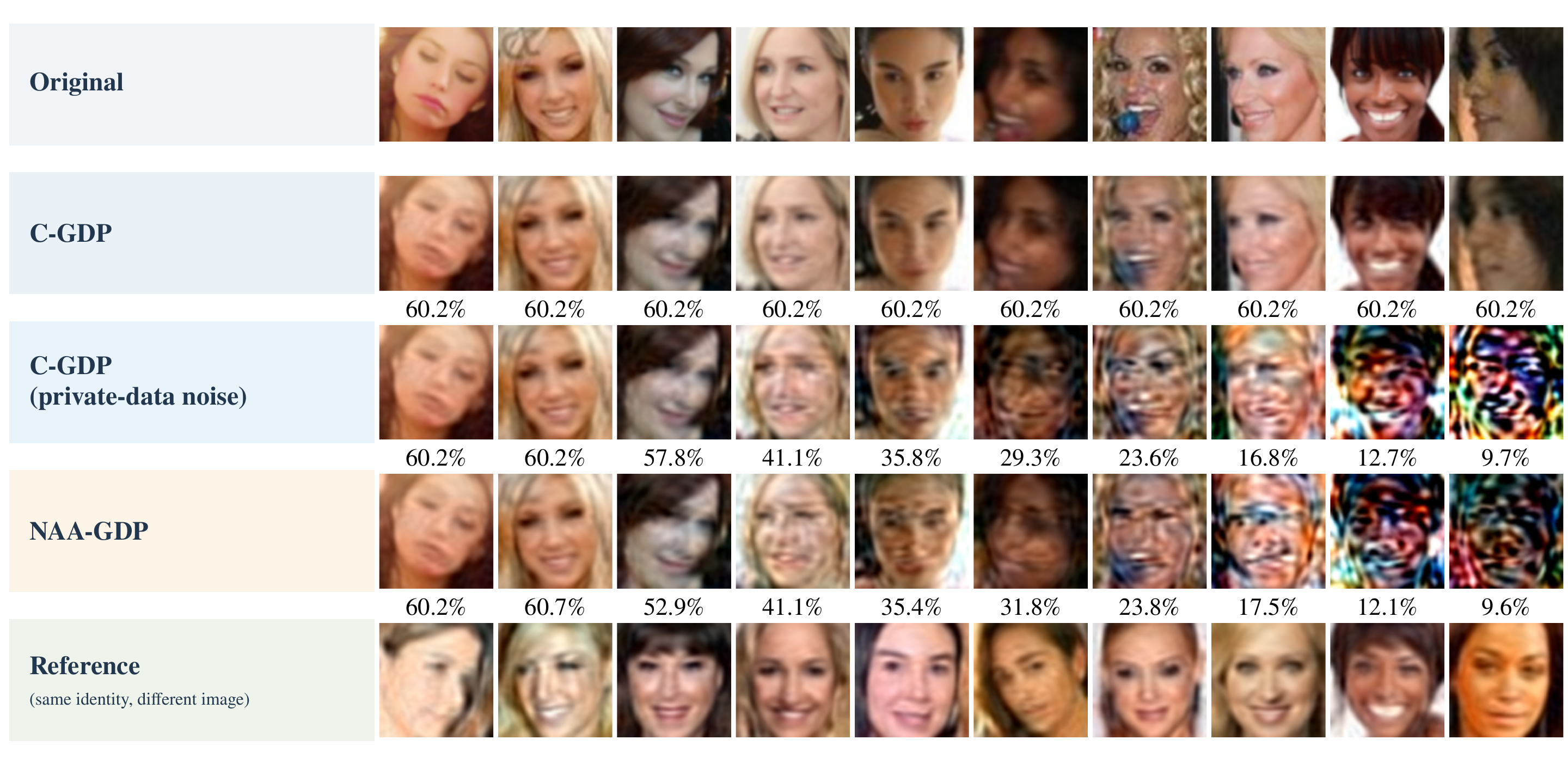}}
\caption{Visual reconstructions under analyst-participant collusion on the CelebA (a) and VGGFace2 (b) datasets. Each column pairs a query with a different reference photograph of the same identity. C-GDP provides the noiseless decoded baseline. Percentages are condition-wide linkage accuracies over 1,000 lineups, not individual-image scores; shared column positions do not imply matched linkage between noise families.}
\label{fig:reconstructions}
\end{figure*}

These results support anchor perturbation as a useful way to reduce measured identity leakage while retaining collaborative utility, with its clearest advantage over private-data noise in the lower-linkage region.

\subsection{Participant-Count Sensitivity}
\label{subsec:participant_sensitivity}

\begin{figure*}[!t]
\centering
\includegraphics[width=\textwidth]{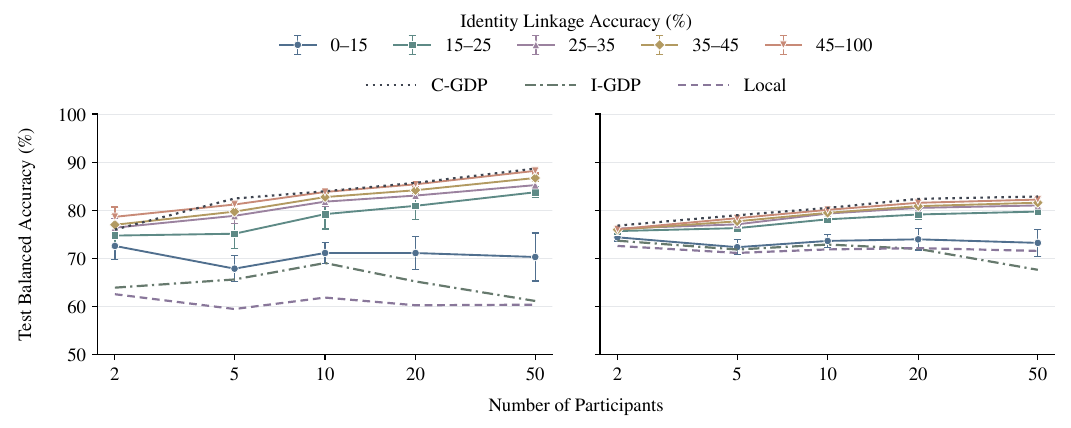}
\caption{Participant-count sensitivity of anchor-noise NAA-GDP on the CelebA (left) and VGGFace2 (right) datasets. Points show mean utility within each linkage interval; error bars show one sample standard deviation.}
\label{fig:participant_count_comparison}
\end{figure*}

Fig.~\ref{fig:participant_count_comparison} examines how collaborative utility changes with the number of participants on the CelebA and VGGFace2 datasets. For each dataset, we use nested collaborations with $p\in\{2,5,10,20,50\}$. Each participant contributes 100 identities and their eligible images, so the collaborations contain between 200 and 5,000 identities. Increasing $p$ therefore expands the available training data and the pooled evaluation cohort. All other settings are consistent with the experiments in Section~\ref{subsec:tradeoff}.

On both datasets, mean NAA-GDP utility increases at every evaluated step from $p=2$ to $p=50$ in the four linkage intervals at or above $15\%$. The gains range from $8.84$ to $9.71$ percentage points on the CelebA dataset and from $4.05$ to $6.16$ percentage points on the VGGFace2 dataset. At $p=50$, balanced accuracy in the highest-linkage interval reaches $88.20\%$ and $82.24\%$, respectively, approaching the C-GDP controls of $88.67\%$ and $82.87\%$ and exceeding I-GDP's $61.18\%$ and $67.65\%$.

In the lowest-linkage interval, $[0,15)\%$, utility is nonmonotonic but remains above Local at every participant count on both datasets. At $p=50$, NAA-GDP achieves $70.31\%$ on the CelebA dataset and $73.24\%$ on the VGGFace2 dataset, compared with Local values of $60.41\%$ and $71.58\%$, respectively. Thus, noisy-anchor alignment translates the additional data from growing collaborations into higher utility, with the scaling benefit depending on measured linkage.

\section{Conclusion}
\label{sec:conclusion}

This paper studied geometric data perturbation for one-shot representation sharing under analyst-participant collusion. We established that, when the centered anchor matrix has full column rank, existing AA-GDP produces the C-GDP representation, while disclosure of the anchor matrix enables exact recovery of participant transformations from noiseless anchor representations. This equivalence also holds when independent isotropic Gaussian noise is added to the private-data representations. Motivated by this limitation, we introduced NAA-GDP, which perturbs the anchor representations while preserving the within-participant geometry of the embedded private data. We formulated alignment as a generalized orthogonal Procrustes problem and used a spectral estimator to recover a shared coordinate system from the noisy anchors.

Experiments on the CelebA and VGGFace2 datasets show that anchor perturbation retains higher mean utility than private-data noise in the lower-linkage region under the evaluated collusion attacks. The participant-count study further shows that noisy-anchor alignment can turn the additional data from growing collaborations into higher utility, with the benefit depending on the measured linkage level. Together, these findings support anchor perturbation as a useful intervention for balancing collaborative utility and resistance to the evaluated reconstruction attacks.

The privacy findings concern the evaluated attacks and fixed data allocations. Preserving exact within-participant distances means that anchor noise alone does not provide record-level differential privacy; appropriately calibrated private-data noise is required when that guarantee is sought~\cite{yamashiro2024dcdp}. Future work includes stronger and adaptive attacks, multiple colluding participants, malicious participants, and evaluation across independent data allocations. Further directions include tighter transformation-estimation bounds, more accurate alignment methods, and nonlinear obfuscation and alignment maps, for which kernel-based Target-Normalized Integration~\cite{suetake2026nonlineardataintegrationkernel} offers a promising foundation.

\bibliographystyle{IEEEtran}
\bibliography{Bibliography}

\end{document}


\maketitle

\begin{abstract}
This supplement complements the main manuscript. Section~I provides the equivalence, recovery, invariance, and max-trace reduction proofs. Section~II presents two additional reconstruction attacks and compares them with the orthogonal-Procrustes estimator on the CelebA and VGGFace2 datasets.
\end{abstract}

\section{Proofs}
\label{sec:supp_proofs}

Definitions and Algorithm~1 refer to the main manuscript, as do the numbers of restated propositions and theorems. Equations below continue the main manuscript's numbering. For the anchor matrix $\bm{A}$ and the anchor representations $\bm{B}_i$, we write
\[
\begin{aligned}
    \bm{a}
    &:=
    \frac{1}{r}\bm{A}^{\top}\mathbf{1}_{r},
    &
    \overline{\bm{A}}
    &:=
    \bm{A}-\mathbf{1}_{r}\bm{a}^{\top},
    \\
    \bm{b}_i
    &:=
    \frac{1}{r}\bm{B}_i^{\top}\mathbf{1}_{r},
    &
    \overline{\bm{B}}_i
    &:=
    \bm{B}_i-\mathbf{1}_{r}\bm{b}_i^{\top}
\end{aligned}
\]
for the column-mean vectors and the centered versions, as in Algorithm~1.

\subsection{Proposition~III.1 (C-GDP Form of AA-GDP)}
\label{subsec:supp_proof_cgdp_form}

\begin{mainprop}{III.1}[C-GDP Form of AA-GDP]
Let $\sigma\geq0$, let every participant $i\in\mathcal{P}$ upload $(\bm{Y}_i,\bm{B}_i)$ as in Definition~II.9, and fix a reference participant $s\in\mathcal{P}$. Assume that $\operatorname{rank}(\overline{\bm{A}})=d$ (which requires $r\geq d+1$). Then the analyst-side alignment of Algorithm~1, applied to these uploads, returns, for every $i\in\mathcal{P}$,
\[
\begin{aligned}
    \bm{R}_i
    &=
    \bm{O}_i^\top\bm{O}_s,
    \\
    \bm{Z}_i
    &=
    \bm{X}_i\bm{O}_s
    +
    \mathbf{1}_{n_i}\Psi_s^\top
    +
    \sigma\bm{W}_i\bm{O}_i^\top\bm{O}_s .
\end{aligned}
\]
Consequently, $\bm{Z}=\begin{bmatrix}\bm{Z}_i\end{bmatrix}_{i\in\mathcal{P}}$ is exactly the upload matrix of C-GDP (private-data noise) with common parameters $(\bm{O}_s,\Psi_s)$ and noise blocks
\[
    \widetilde{\bm{W}}_i
    :=
    \bm{W}_i\bm{O}_i^\top\bm{O}_s,
    \qquad i\in\mathcal{P}.
\]
Moreover, the entries of these blocks are jointly i.i.d.\ $\mathcal{N}(0,1)$ and independent of all GDP parameters.
\end{mainprop}

The proof follows the three analyst-side steps of Algorithm~1: two lemmas establish the centering identities and the relative rotations, and the proof of the proposition applies the alignment maps. Private-data noise enters only in the last step, because the anchor representations of Definition~II.9 are noiseless for every $\sigma\geq0$.

\begin{lemma}[Centering Identities]
\label{lem:supp_aaigdp_centering}
Let participant $i\in\mathcal{P}$ upload the anchor representation $\bm{B}_i=\bm{A}\bm{O}_i+\mathbf{1}_{r}\Psi_i^\top$ of Definition~II.9. Then
\begin{equation}
    \bm{b}_{i}
    =
    \bm{O}_i^\top\bm{a}
    +
    \Psi_i,
    \qquad
    \overline{\bm{B}}_i
    =
    \overline{\bm{A}}\bm{O}_i .
    \label{eq:supp_centering_identities}
\end{equation}
Moreover, $\mathbf{1}_r^\top\overline{\bm{A}}=\bm{0}_d^\top$, and hence $\operatorname{rank}(\overline{\bm{A}})\leq r-1$.
\end{lemma}

\begin{proof}
By the definition of $\bm{b}_i$ and the expression for $\bm{B}_i$,
\begin{equation}
\begin{aligned}
    \bm{b}_{i}
    &=
    \frac{1}{r}\bm{B}_i^\top\mathbf{1}_{r} \\
    &=
    \frac{1}{r}
    \left(
        \bm{O}_i^\top\bm{A}^\top
        +
        \Psi_i\mathbf{1}_{r}^\top
    \right)
    \mathbf{1}_{r} \\
    &=
    \bm{O}_i^\top
    \left(
        \frac{1}{r}\bm{A}^\top\mathbf{1}_{r}
    \right)
    +
    \Psi_i \\
    &=
    \bm{O}_i^\top\bm{a}
    +
    \Psi_i .
\end{aligned}
\end{equation}
Equivalently, $\bm{b}_{i}^{\top}=\bm{a}^{\top}\bm{O}_i+\Psi_i^\top$. Therefore,
\begin{equation}
\begin{aligned}
    \overline{\bm{B}}_i
    &=
    \bm{B}_i
    -
    \mathbf{1}_{r}\bm{b}_{i}^{\top} \\
    &=
    \bm{A}\bm{O}_i
    +
    \mathbf{1}_{r}\Psi_i^\top
    -
    \mathbf{1}_{r}
    \left(
        \bm{a}^{\top}\bm{O}_i
        +
        \Psi_i^\top
    \right) \\
    &=
    \left(
        \bm{A}
        -
        \mathbf{1}_{r}\bm{a}^{\top}
    \right)
    \bm{O}_i \\
    &=
    \overline{\bm{A}}\bm{O}_i .
\end{aligned}
\end{equation}
Finally, the definition of $\bm a$ gives
$\mathbf{1}_r^\top\overline{\bm A}
=\mathbf{1}_r^\top\bm A-r\bm a^\top
=\bm 0_d^\top$.
Thus, the rows of $\overline{\bm A}$ are linearly dependent, so
$\operatorname{rank}(\overline{\bm A})\leq r-1$.
\end{proof}

\begin{lemma}[Relative Rotations]
\label{lem:supp_aaigdp_relative_rotation}
Let every participant $i\in\mathcal{P}$ upload as in Definition~II.9, assume $\operatorname{rank}(\overline{\bm A})=d$, and fix the reference participant $s\in\mathcal P$. Then $r\geq d+1$, and the rotations $\bm R_i=\bm U_i\bm V_i^\top$ computed by Algorithm~1 from the SVD
\begin{equation}
    \overline{\bm{B}}_i^\top \overline{\bm{B}}_s
    =
    \bm{U}_i\bm{\Sigma}_i\bm{V}_i^\top
    \label{eq:supp_relative_rotation_svd}
\end{equation}
satisfy
\begin{equation}
    \bm{R}_i
    =
    \bm{O}_i^\top\bm{O}_s
    \qquad
    \text{for every }i\in\mathcal P .
    \label{eq:supp_relative_rotation}
\end{equation}
\end{lemma}

\begin{proof}
Lemma~\ref{lem:supp_aaigdp_centering} gives
$\operatorname{rank}(\overline{\bm A})\leq r-1$.
Full column rank therefore implies $d\leq r-1$, equivalently $r\geq d+1$.
The same lemma also gives
\begin{equation}
    \overline{\bm{B}}_i
    =
    \overline{\bm{A}}\bm{O}_i,
    \qquad
    \overline{\bm{B}}_s
    =
    \overline{\bm{A}}\bm{O}_s .
\end{equation}
Consequently,
\begin{equation}
    \overline{\bm{B}}_i^\top \overline{\bm{B}}_s
    =
    \bm{O}_i^\top
    \overline{\bm{A}}^\top\overline{\bm{A}}
    \bm{O}_s .
\end{equation}
Since $\overline{\bm{A}}$ has full column rank,
$\overline{\bm{A}}^\top\overline{\bm{A}}$ is symmetric positive definite, as is
$\bm{O}_s^\top\overline{\bm{A}}^\top\overline{\bm{A}}\bm{O}_s$.
For every $i\in\mathcal{P}$, the factorization
\begin{equation}
\overline{\bm{B}}_i^\top \overline{\bm{B}}_s
=
\left(\bm{O}_i^\top\bm{O}_s\right)
\left(
\bm{O}_s^\top
\overline{\bm{A}}^\top\overline{\bm{A}}
\bm{O}_s
\right)
\end{equation}
is therefore the polar decomposition of
$\overline{\bm{B}}_i^\top\overline{\bm{B}}_s$ (for $i=s$, with orthogonal factor $\bm{O}_s^\top\bm{O}_s=\bm{I}_d$).
This matrix is nonsingular, so its orthogonal polar factor is unique.
Its SVD~\eqref{eq:supp_relative_rotation_svd} gives the same polar factor, and hence
\begin{equation}
    \bm{U}_i\bm{V}_i^\top
    =
    \bm{O}_i^\top\bm{O}_s ,
\end{equation}
which is \eqref{eq:supp_relative_rotation} by the definition of $\bm R_i$ in Algorithm~1.
\end{proof}

\begin{proof}[Proof of Proposition~III.1]
Lemma~\ref{lem:supp_aaigdp_relative_rotation} gives $\bm{R}_i=\bm{O}_i^\top\bm{O}_s$ for every $i\in\mathcal{P}$, which is the first identity. For the second, fix $i\in\mathcal{P}$. By Lemma~\ref{lem:supp_aaigdp_centering}, $\bm b_i^\top=\bm a^\top\bm O_i+\Psi_i^\top$ and $\bm b_s^\top=\bm a^\top\bm O_s+\Psi_s^\top$. Substituting these, $\bm R_i=\bm O_i^\top\bm O_s$, and the private-data representation $\bm{Y}_i=\bm{X}_i\bm{O}_i+\mathbf{1}_{n_i}\Psi_i^\top+\sigma\bm{W}_i$ of Definition~II.9 into the alignment map of Algorithm~1 yields
\begin{equation}
\begin{aligned}
\bm{Z}_i
&=
\mathcal{N}_i(\bm{Y}_i) \\
&=
\left(
\bm{X}_i\bm{O}_i
+\mathbf{1}_{n_i}\Psi_i^\top
+\sigma\bm{W}_i
-\mathbf{1}_{n_i}\bm{b}_{i}^{\top}
\right)\bm{R}_i \\
&\quad
+\mathbf{1}_{n_i}\bm{b}_{s}^{\top} \\
&=
\left(
\bm{X}_i\bm{O}_i
-\mathbf{1}_{n_i}\bm{a}^\top\bm{O}_i
+\sigma\bm{W}_i
\right)
\bm{O}_i^\top\bm{O}_s \\
&\quad
+\mathbf{1}_{n_i}
\left(
\bm{a}^\top\bm{O}_s+\Psi_s^\top
\right) \\
&=
\left(
\bm{X}_i-\mathbf{1}_{n_i}\bm{a}^\top
\right)\bm{O}_s
+\sigma\bm{W}_i\bm{O}_i^\top\bm{O}_s \\
&\quad
+\mathbf{1}_{n_i}\bm{a}^\top\bm{O}_s
+\mathbf{1}_{n_i}\Psi_s^\top \\
&=
\bm{X}_i\bm{O}_s
+\mathbf{1}_{n_i}\Psi_s^\top
+\sigma\bm{W}_i\bm{O}_i^\top\bm{O}_s ,
\end{aligned}
\label{eq:supp_noisy_alignment_identity}
\end{equation}
which is the second identity. Stacking the blocks, with $\widetilde{\bm{W}}_i=\bm{W}_i\bm{O}_i^\top\bm{O}_s$ as in the proposition,
\begin{equation}
\begin{aligned}
    \bm{Z}
    =
    \begin{bmatrix}
        \bm{Z}_i
    \end{bmatrix}_{i\in\mathcal{P}}
    &=
    \begin{bmatrix}
        \bm{X}_i\bm{O}_s
        +
        \mathbf{1}_{n_i}\Psi_s^\top
        +
        \sigma\widetilde{\bm{W}}_i
    \end{bmatrix}_{i\in\mathcal{P}} \\
    &=
    \bm{X}\bm{O}_s
    +
    \mathbf{1}_{n}\Psi_s^\top
    +
    \sigma
    \begin{bmatrix}
        \widetilde{\bm{W}}_i
    \end{bmatrix}_{i\in\mathcal{P}},
\end{aligned}
\label{eq:supp_cgdp_form_stacked}
\end{equation}
which is the upload matrix of C-GDP (private-data noise) of Definition~II.5 with common parameters $(\bm{O}_s,\Psi_s)$ and noise matrices $\widetilde{\bm{W}}_1,\ldots,\widetilde{\bm{W}}_p$.

Finally, since the matrices $\bm{W}_1,\ldots,\bm{W}_p$ are independent of one another and of the GDP parameters and have i.i.d.\ $\mathcal{N}(0,1)$ entries (Definition~II.9), and since the standard Gaussian distribution is invariant under orthogonal transformations, the blocks $\widetilde{\bm{W}}_i=\bm{W}_i\bm{O}_i^\top\bm{O}_s$ have jointly i.i.d.\ $\mathcal{N}(0,1)$ entries for every value of the GDP parameters and are therefore independent of them.
\end{proof}

\subsection{Theorem~III.2 (Known-Anchor Recovery under Collusion)}
\label{subsec:supp_proof_known_anchor_recovery}

\begin{mainthm}{III.2}[Known-Anchor Recovery under Collusion]
Consider the analyst-participant collusion setting of Definition~II.10, let every participant $i\in\mathcal{P}$ upload $(\bm{Y}_i,\bm{B}_i)$ as in Definition~II.9 with $\sigma\geq0$, and assume $\operatorname{rank}(\overline{\bm{A}})=d$. Then, for every non-colluding participant $i\in\mathcal{P}\setminus\{c\}$, the attacker recovers the transformation parameters $(\bm{O}_i,\Psi_i)$ exactly from $(\bm{A},\bm{B}_i)$ and inverts the upload as
\[
    \left(
        \bm{Y}_i
        -
        \mathbf{1}_{n_i}\Psi_i^\top
    \right)
    \bm{O}_i^\top
    =
    \bm{X}_i
    +
    \sigma\bm{W}_i\bm{O}_i^\top.
\]
\end{mainthm}

\begin{proof}
By Definition~II.10, the attacker holds the anchor matrix $\bm{A}$ disclosed by the colluding participant and every uploaded pair $(\bm{Y}_i,\bm{B}_i)$, hence also $\bm{a}$, $\overline{\bm{A}}$, $\bm{b}_i$, and $\overline{\bm{B}}_i$ for every $i\in\mathcal{P}\setminus\{c\}$. The anchor representation $\bm{B}_i$ of Definition~II.9 is noiseless for every $\sigma\geq0$, so Lemma~\ref{lem:supp_aaigdp_centering} gives $\overline{\bm{B}}_i=\overline{\bm{A}}\bm{O}_i$ and $\bm{b}_i=\bm{O}_i^\top\bm{a}+\Psi_i$, and $\operatorname{rank}(\overline{\bm{A}})=d$ gives $\overline{\bm{A}}^{\dagger}\overline{\bm{A}}=\bm{I}_d$. Therefore,
\begin{equation}
    \overline{\bm{A}}^{\dagger}\overline{\bm{B}}_i
    =
    \overline{\bm{A}}^{\dagger}\overline{\bm{A}}\bm{O}_i
    =
    \bm{O}_i,
    \qquad
    \bm{b}_i-\bm{O}_i^\top\bm{a}
    =
    \Psi_i ,
    \label{eq:supp_known_anchor_recovery}
\end{equation}
and since both left-hand sides are computed from $(\bm{A},\bm{B}_i)$ alone, the attacker recovers $(\bm{O}_i,\Psi_i)$ exactly. Substituting the private-data representation $\bm{Y}_i=\bm{X}_i\bm{O}_i+\mathbf{1}_{n_i}\Psi_i^\top+\sigma\bm{W}_i$ of Definition~II.9 and using $\bm{O}_i\bm{O}_i^\top=\bm{I}_d$ then yields
\begin{equation}
\begin{aligned}
    \left(
        \bm{Y}_i
        -
        \mathbf{1}_{n_i}\Psi_i^\top
    \right)
    \bm{O}_i^\top
    &=
    \left(
        \bm{X}_i\bm{O}_i
        +
        \sigma\bm{W}_i
    \right)
    \bm{O}_i^\top \\
    &=
    \bm{X}_i
    +
    \sigma\bm{W}_i\bm{O}_i^\top ,
\end{aligned}
\label{eq:supp_known_anchor_inversion}
\end{equation}
which is the inversion identity of the theorem.
\end{proof}

\subsection{Proposition~IV.2 (Rigid-Motion Invariance)}
\label{subsec:supp_proof_gopp_invariance}

Propositions~IV.2 and~IV.3 concern the following alignment objective of the main manuscript, formulated over a template $\bm{C}\in\mathbb{R}^{r\times d}$, rotations $\bm{R}_i\in\mathcal{O}(d)$, and translations $\boldsymbol{\tau}_i\in\mathbb{R}^{d}$ for the anchor representations $\bm{B}_i$ of Definition~IV.1:
\begin{equation}
\label{prob:supp_gopp_full}
\begin{aligned}
\min_{\substack{
    \bm{C},\,\bm{R}_1,\ldots,\bm{R}_p,\\
    \boldsymbol{\tau}_1,\ldots,\boldsymbol{\tau}_p}}
&\quad
\sum_{i=1}^{p}
\left\|
        \bm{C}
        -
        \left(
            \bm{B}_i
            -
            \mathbf{1}_{r}\boldsymbol{\tau}_i^\top
        \right)
        \bm{R}_i
    \right\|_F^2  \\
\text{s.t.}
&\quad
\bm{R}_i\in\mathcal{O}(d),
\quad
i\in\mathcal{P}.
\end{aligned}
\end{equation}

\begin{mainprop}{IV.2}[Rigid-Motion Invariance]
Let \((\bm{C},\bm{R}_1,\ldots,\bm{R}_p,\boldsymbol{\tau}_1,\ldots,\boldsymbol{\tau}_p)\) be feasible for Problem~\eqref{prob:supp_gopp_full}. For any \(\boldsymbol{s}\in\mathbb{R}^{d}\) and \(\bm{Q}\in\mathcal{O}(d)\), define
\[
\begin{aligned}
    \widetilde{\bm{C}}
    &=
    \left(\bm{C}+\mathbf{1}_{r}\boldsymbol{s}^\top\right)\bm{Q},
    \\
    \widetilde{\bm{R}}_i
    &=
    \bm{R}_i\bm{Q},
    \qquad
    \widetilde{\boldsymbol{\tau}}_i
    =
    \boldsymbol{\tau}_i
    -
    \bm{R}_i\boldsymbol{s},
    \qquad
    i\in\mathcal{P}.
\end{aligned}
\]
Then \((\widetilde{\bm{C}},\widetilde{\bm{R}}_1,\ldots,\widetilde{\bm{R}}_p,\widetilde{\boldsymbol{\tau}}_1,\ldots,\widetilde{\boldsymbol{\tau}}_p)\) is feasible and attains the same objective value.
\end{mainprop}

\begin{proof}
Since $\bm{R}_i$ and $\bm{Q}$ are orthogonal, so is $\widetilde{\bm{R}}_i=\bm{R}_i\bm{Q}$, and the transformed tuple is feasible. For its objective value, substituting the transformed variables into the $i$-th summand of \eqref{prob:supp_gopp_full} and using $\bm{R}_i^\top\bm{R}_i=\bm{I}_d$ gives
\begin{equation}
\begin{aligned}
&\widetilde{\bm{C}}
-
\left(
    \bm{B}_i-\mathbf{1}_{r}\widetilde{\boldsymbol{\tau}}_i^\top
\right)
\widetilde{\bm{R}}_i
\\
&\quad=
\left(\bm{C}+\mathbf{1}_{r}\boldsymbol{s}^\top\right)\bm{Q}
-
\left(
    \bm{B}_i-\mathbf{1}_{r}\boldsymbol{\tau}_i^\top
    +\mathbf{1}_{r}\boldsymbol{s}^\top\bm{R}_i^\top
\right)
\bm{R}_i\bm{Q}
\\
&\quad=
\left[
    \bm{C}
    -
    \left(
        \bm{B}_i-\mathbf{1}_{r}\boldsymbol{\tau}_i^\top
    \right)
    \bm{R}_i
\right]\bm{Q}
\\
&\qquad\quad
+
\mathbf{1}_{r}\boldsymbol{s}^\top\bm{Q}
-
\mathbf{1}_{r}\boldsymbol{s}^\top\bm{R}_i^\top\bm{R}_i\bm{Q}
\\
&\quad=
\left[
    \bm{C}
    -
    \left(
        \bm{B}_i-\mathbf{1}_{r}\boldsymbol{\tau}_i^\top
    \right)
    \bm{R}_i
\right]\bm{Q} .
\end{aligned}
\label{eq:supp_gopp_invariance_summand}
\end{equation}
Right multiplication by the orthogonal matrix $\bm{Q}$ preserves the Frobenius norm, so every summand, and hence the objective value, is unchanged.
\end{proof}

\subsection{Proposition~IV.3 (Max-Trace Reduction)}
\label{subsec:supp_proof_gopp_reduction}

\begin{mainprop}{IV.3}[Max-Trace Reduction]
Consider Problem~\eqref{prob:supp_gopp_full} with the additional constraint \(\mathbf{1}_r^\top\bm{C}=\mathbf{0}_d^\top\). Every minimizer \((\bm{C}^\star,\bm{R}_1^\star,\ldots,\bm{R}_p^\star,\boldsymbol{\tau}_1^\star,\ldots,\boldsymbol{\tau}_p^\star)\) of the constrained problem satisfies
\[
    \boldsymbol{\tau}_i^\star
    =
    \bm{b}_i,
    \quad
    i\in\mathcal{P},
    \qquad
    \bm{C}^\star
    =
    \frac{1}{p}
    \sum_{i=1}^{p}
    \overline{\bm{B}}_i\bm{R}_i^\star,
\]
and \((\bm{R}_1^\star,\ldots,\bm{R}_p^\star)\) solves
\begin{equation}
    \max_{\bm{R}_1,\ldots,\bm{R}_p\in\mathcal{O}(d)}
    \operatorname{tr}
    \left(
        \bm{R}^\top\bm{G}\bm{R}
    \right),
    \label{prob:supp_gopp_trace_form}
\end{equation}
where \(\bm{R}:=\begin{bmatrix}\bm{R}_i\end{bmatrix}_{i\in[p]}\in\mathbb{R}^{pd\times d}\) stacks the rotations and \(\bm{G}\in\mathbb{R}^{pd\times pd}\) is the block matrix with \((i,j)\) block \(\overline{\bm{B}}_i^\top\overline{\bm{B}}_j\). Conversely, every maximizer of \eqref{prob:supp_gopp_trace_form}, together with \(\boldsymbol{\tau}_i^\star\) and \(\bm{C}^\star\) as above, minimizes the constrained problem.
\end{mainprop}

The proof eliminates the translations and the template in turn: the first two lemmas minimize the objective of \eqref{prob:supp_gopp_full} over the translations and over the template with the remaining variables fixed, the third evaluates the resulting objective, and the proof of the proposition combines them.

\begin{lemma}[Optimal Translations]
\label{lem:supp_gopp_translations}
Let $\bm{C}\in\mathbb{R}^{r\times d}$ satisfy $\mathbf{1}_r^\top\bm{C}=\mathbf{0}_d^\top$, and let $\bm{R}_1,\ldots,\bm{R}_p\in\mathcal{O}(d)$. Then, for all $\boldsymbol{\tau}_1,\ldots,\boldsymbol{\tau}_p\in\mathbb{R}^{d}$,
\begin{equation}
\begin{aligned}
    &\sum_{i=1}^{p}
    \left\|
        \bm{C}
        -
        \left(
            \bm{B}_i-\mathbf{1}_{r}\boldsymbol{\tau}_i^\top
        \right)
        \bm{R}_i
    \right\|_F^2
    \\
    &\quad\geq
    \sum_{i=1}^{p}
    \left\|
        \bm{C}
        -
        \overline{\bm{B}}_i\bm{R}_i
    \right\|_F^2 ,
\end{aligned}
\label{eq:supp_gopp_translation_bound}
\end{equation}
with equality if and only if $\boldsymbol{\tau}_i=\bm{b}_i$ for every $i\in\mathcal{P}$.
\end{lemma}

\begin{proof}
Fix $i\in\mathcal{P}$. Because right multiplication by $\bm{R}_i^\top$ preserves the Frobenius norm,
\begin{equation}
\begin{aligned}
    &\left\|
        \bm{C}
        -
        \left(
            \bm{B}_i-\mathbf{1}_{r}\boldsymbol{\tau}_i^\top
        \right)
        \bm{R}_i
    \right\|_F
    \\
    &\quad=
    \left\|
        \left(
            \bm{B}_i-\bm{C}\bm{R}_i^\top
        \right)
        -
        \mathbf{1}_{r}\boldsymbol{\tau}_i^\top
    \right\|_F ,
\end{aligned}
\label{eq:supp_gopp_summand_rewritten}
\end{equation}
and the right-hand side is an ordinary least-squares problem in $\boldsymbol{\tau}_i$, whose unique minimizer is the column mean of $\bm{B}_i-\bm{C}\bm{R}_i^\top$,
\begin{equation}
    \frac{1}{r}
    \left(
        \bm{B}_i-\bm{C}\bm{R}_i^\top
    \right)^\top
    \mathbf{1}_{r}
    =
    \bm{b}_i
    -
    \frac{1}{r}\bm{R}_i\bm{C}^\top\mathbf{1}_{r}
    =
    \bm{b}_i ,
    \label{eq:supp_gopp_translation}
\end{equation}
by $\mathbf{1}_r^\top\bm{C}=\mathbf{0}_d^\top$. Hence, the $i$-th summand on the left of \eqref{eq:supp_gopp_translation_bound} is at least its value at $\boldsymbol{\tau}_i=\bm{b}_i$, with equality if and only if $\boldsymbol{\tau}_i=\bm{b}_i$. Since $\bm{B}_i-\mathbf{1}_{r}\bm{b}_i^\top=\overline{\bm{B}}_i$, that value is $\|\bm{C}-\overline{\bm{B}}_i\bm{R}_i\|_F^2$. Summing over $i\in\mathcal{P}$ proves the claim.
\end{proof}

\begin{lemma}[Optimal Template]
\label{lem:supp_gopp_template}
Let $\bm{R}_1,\ldots,\bm{R}_p\in\mathcal{O}(d)$ and define
\begin{equation}
    \bm{C}_{\bm{R}}
    :=
    \frac{1}{p}
    \sum_{i=1}^{p}
    \overline{\bm{B}}_i\bm{R}_i .
    \label{eq:supp_gopp_template}
\end{equation}
Then $\mathbf{1}_r^\top\bm{C}_{\bm{R}}=\mathbf{0}_d^\top$, and every $\bm{C}\in\mathbb{R}^{r\times d}$ satisfies
\begin{equation}
\begin{aligned}
    \sum_{i=1}^{p}
    \left\|
        \bm{C}
        -
        \overline{\bm{B}}_i\bm{R}_i
    \right\|_F^2
    &=
    \sum_{i=1}^{p}
    \left\|
        \bm{C}_{\bm{R}}
        -
        \overline{\bm{B}}_i\bm{R}_i
    \right\|_F^2
    \\
    &\quad
    +
    p
    \left\|
        \bm{C}-\bm{C}_{\bm{R}}
    \right\|_F^2 .
\end{aligned}
\label{eq:supp_gopp_template_identity}
\end{equation}
In particular, $\bm{C}_{\bm{R}}$ is the unique minimizer of the left-hand side.
\end{lemma}

\begin{proof}
The template is centered because every $\overline{\bm{B}}_i$ is centered:
\begin{equation}
    \mathbf{1}_r^\top\bm{C}_{\bm{R}}
    =
    \frac{1}{p}
    \sum_{i=1}^{p}
    \mathbf{1}_r^\top\overline{\bm{B}}_i\bm{R}_i
    =
    \mathbf{0}_d^\top .
    \label{eq:supp_gopp_template_centered}
\end{equation}
For the identity, let $\bm{D}_i:=\bm{C}_{\bm{R}}-\overline{\bm{B}}_i\bm{R}_i$ for $i\in\mathcal{P}$. By \eqref{eq:supp_gopp_template},
\begin{equation}
    \sum_{i=1}^{p}\bm{D}_i
    =
    p\bm{C}_{\bm{R}}
    -
    \sum_{i=1}^{p}
    \overline{\bm{B}}_i\bm{R}_i
    =
    \bm{0} .
    \label{eq:supp_gopp_deviations_sum}
\end{equation}
Writing $\bm{C}-\overline{\bm{B}}_i\bm{R}_i=(\bm{C}-\bm{C}_{\bm{R}})+\bm{D}_i$ and expanding the squared Frobenius norm,
\begin{equation}
\begin{aligned}
    \sum_{i=1}^{p}
    \left\|
        \bm{C}
        -
        \overline{\bm{B}}_i\bm{R}_i
    \right\|_F^2
    &=
    p
    \left\|
        \bm{C}-\bm{C}_{\bm{R}}
    \right\|_F^2
    \\
    &\quad
    +
    2\operatorname{tr}
    \Bigl(
        \left(\bm{C}-\bm{C}_{\bm{R}}\right)^\top
        \sum_{i=1}^{p}\bm{D}_i
    \Bigr)
    \\
    &\quad
    +
    \sum_{i=1}^{p}
    \left\|
        \bm{D}_i
    \right\|_F^2 ,
\end{aligned}
\label{eq:supp_gopp_template_expansion}
\end{equation}
and the middle term vanishes by \eqref{eq:supp_gopp_deviations_sum}, which proves \eqref{eq:supp_gopp_template_identity}. Since $p\|\bm{C}-\bm{C}_{\bm{R}}\|_F^2$ is positive unless $\bm{C}=\bm{C}_{\bm{R}}$, the minimizer is unique.
\end{proof}

\begin{lemma}[Reduced Objective]
\label{lem:supp_gopp_reduced_objective}
Let $\bm{R}_1,\ldots,\bm{R}_p\in\mathcal{O}(d)$, let $\bm{C}_{\bm{R}}$ be as in \eqref{eq:supp_gopp_template}, and let $\bm{R}$ and $\bm{G}$ be as in Proposition~IV.3. Then
\begin{equation}
    \sum_{i=1}^{p}
    \left\|
        \bm{C}_{\bm{R}}
        -
        \overline{\bm{B}}_i\bm{R}_i
    \right\|_F^2
    =
    \sum_{i=1}^{p}
    \left\|
        \overline{\bm{B}}_i
    \right\|_F^2
    -
    \frac{1}{p}
    \operatorname{tr}
    \left(
        \bm{R}^\top\bm{G}\bm{R}
    \right).
    \label{eq:supp_gopp_reduced_objective}
\end{equation}
\end{lemma}

\begin{proof}
Expanding the squares, using $\sum_{i=1}^{p}\overline{\bm{B}}_i\bm{R}_i=p\bm{C}_{\bm{R}}$ for the cross term and $\|\overline{\bm{B}}_i\bm{R}_i\|_F=\|\overline{\bm{B}}_i\|_F$,
\begin{equation}
\begin{aligned}
    &\sum_{i=1}^{p}
    \left\|
        \bm{C}_{\bm{R}}
        -
        \overline{\bm{B}}_i\bm{R}_i
    \right\|_F^2
    \\
    &\quad=
    p
    \left\|
        \bm{C}_{\bm{R}}
    \right\|_F^2
    -
    2p
    \left\|
        \bm{C}_{\bm{R}}
    \right\|_F^2
    +
    \sum_{i=1}^{p}
    \left\|
        \overline{\bm{B}}_i\bm{R}_i
    \right\|_F^2
    \\
    &\quad=
    \sum_{i=1}^{p}
    \left\|
        \overline{\bm{B}}_i
    \right\|_F^2
    -
    p
    \left\|
        \bm{C}_{\bm{R}}
    \right\|_F^2 .
\end{aligned}
\label{eq:supp_gopp_reduced_expansion}
\end{equation}
Finally, because the $(i,j)$ block of $\bm{G}$ is $\overline{\bm{B}}_i^\top\overline{\bm{B}}_j$,
\begin{equation}
\begin{aligned}
    p
    \left\|
        \bm{C}_{\bm{R}}
    \right\|_F^2
    &=
    \frac{1}{p}
    \left\|
        \sum_{i=1}^{p}
        \overline{\bm{B}}_i\bm{R}_i
    \right\|_F^2
    \\
    &=
    \frac{1}{p}
    \sum_{i=1}^{p}
    \sum_{j=1}^{p}
    \operatorname{tr}
    \left(
        \bm{R}_i^\top
        \overline{\bm{B}}_i^\top
        \overline{\bm{B}}_j
        \bm{R}_j
    \right)
    \\
    &=
    \frac{1}{p}
    \operatorname{tr}
    \left(
        \bm{R}^\top\bm{G}\bm{R}
    \right).
\end{aligned}
\label{eq:supp_gopp_trace_identity}
\end{equation}
\end{proof}

\begin{proof}[Proof of Proposition~IV.3]
Fix a feasible point of the constrained problem, with template $\bm{C}$, rotations $\bm{R}_1,\ldots,\bm{R}_p$, and translations $\boldsymbol{\tau}_1,\ldots,\boldsymbol{\tau}_p$. Lemmas~\ref{lem:supp_gopp_translations}, \ref{lem:supp_gopp_template}, and~\ref{lem:supp_gopp_reduced_objective} give
\begin{equation}
\begin{aligned}
    &\sum_{i=1}^{p}
    \left\|
        \bm{C}
        -
        \left(
            \bm{B}_i-\mathbf{1}_{r}\boldsymbol{\tau}_i^\top
        \right)
        \bm{R}_i
    \right\|_F^2
    \\
    &\quad\geq
    \sum_{i=1}^{p}
    \left\|
        \bm{C}_{\bm{R}}
        -
        \overline{\bm{B}}_i\bm{R}_i
    \right\|_F^2
    \\
    &\quad=
    \sum_{i=1}^{p}
    \left\|
        \overline{\bm{B}}_i
    \right\|_F^2
    -
    \frac{1}{p}
    \operatorname{tr}
    \left(
        \bm{R}^\top\bm{G}\bm{R}
    \right),
\end{aligned}
\label{eq:supp_gopp_lower_bound}
\end{equation}
with equality if and only if $\boldsymbol{\tau}_i=\bm{b}_i$ for every $i\in\mathcal{P}$ and $\bm{C}=\bm{C}_{\bm{R}}$.

The right-hand side of \eqref{eq:supp_gopp_lower_bound} depends on the rotations only through $\operatorname{tr}(\bm{R}^\top\bm{G}\bm{R})$. Its smallest value over $\bm{R}_1,\ldots,\bm{R}_p\in\mathcal{O}(d)$ is therefore attained exactly at the maximizers of \eqref{prob:supp_gopp_trace_form}, which exist because $\mathcal{O}(d)$ is compact. Each maximizer, completed by $\boldsymbol{\tau}_i=\bm{b}_i$ and $\bm{C}=\bm{C}_{\bm{R}}$, is feasible for the constrained problem by Lemma~\ref{lem:supp_gopp_template} and attains equality in \eqref{eq:supp_gopp_lower_bound}. Hence, the smallest value of the right-hand side is the minimum of the constrained problem.

Every minimizer attains this minimum, so equality holds in \eqref{eq:supp_gopp_lower_bound} and its rotations maximize \eqref{prob:supp_gopp_trace_form}; that is, $\boldsymbol{\tau}_i^\star=\bm{b}_i$ for every $i\in\mathcal{P}$, $\bm{C}^\star=\bm{C}_{\bm{R}^\star}$, and $(\bm{R}_1^\star,\ldots,\bm{R}_p^\star)$ solves \eqref{prob:supp_gopp_trace_form}. Conversely, every maximizer of \eqref{prob:supp_gopp_trace_form} completed in this way attains the minimum and is therefore a minimizer.
\end{proof}

\section{Additional Reconstruction Attacks}
\label{sec:supp_attacks}

We define the Moore--Penrose (MP) and alignment-map calibration (AM) estimators as alternatives to the orthogonal-Procrustes (OP) estimator used in the main manuscript. Under the same NAA-GDP analyst--participant collusion model, the attacker observes the uploads $(\bm Y_j,\bm B_j)$ and spectral alignment rotations $\widehat{\bm R}_j$; collusion with participant $c$ also discloses the shared anchor matrix $\bm A$ and the colluder's transformation parameters. We consider a protected participant $i\ne c$, with $\mathbf 1_r^\top\bm A=\mathbf 0_d^\top$ and $\bm A^\top\bm A=\bm I_d$.

Using the row-vector convention of the main manuscript, the target uploads are
\begin{equation}
    \begin{aligned}
    \bm Y_i&=\bm X_i\bm O_i+\mathbf 1_{n_i}\Psi_i^\top,\\
    \bm B_i&=\bm A\bm O_i+\mathbf 1_r\Psi_i^\top+v\bm W_i,
    \end{aligned}
    \label{eq:supp_attack_uploads}
\end{equation}
where $\bm O_i\in\mathcal O(d)$ and the entries of $\bm W_i$ are independent standard Gaussian variables. Define
\begin{equation}
    \bm b_i:=\frac{1}{r}\bm B_i^\top\mathbf 1_r,
    \qquad
    \overline{\bm B}_i:=\bm B_i-\mathbf 1_r\bm b_i^\top.
    \label{eq:supp_attack_centering}
\end{equation}
Writing $\bm H_r=\bm I_r-r^{-1}\mathbf 1_r\mathbf 1_r^\top$ gives
\begin{equation}
    \begin{aligned}
    \bm b_i&=\Psi_i+\boldsymbol\eta_i,
    \quad
    \boldsymbol\eta_i:=\frac{v}{r}\bm W_i^\top\mathbf 1_r,
    \\
    \overline{\bm B}_i&=\bm A\bm O_i+v\bm H_r\bm W_i.
    \end{aligned}
    \label{eq:supp_attack_noisy_relations}
\end{equation}
Each estimator uses $\widehat\Psi_i=\bm b_i$. Subtracting this estimate leaves the anchor-noise mean in the reconstruction, even though the private-data upload is unnoised:
\begin{equation}
    \begin{aligned}
    \widehat{\bm X}_i
    &=\left(\bm Y_i-\mathbf 1_{n_i}\bm b_i^\top\right)
       \widehat{\bm O}_i^\dagger\\
    &=\left(\bm X_i\bm O_i-\mathbf 1_{n_i}\boldsymbol\eta_i^\top\right)
       \widehat{\bm O}_i^\dagger.
    \end{aligned}
    \label{eq:supp_attack_reconstruction}
\end{equation}
For an orthogonal estimate, the pseudoinverse is its transpose.

When GDP acts on embedded rows $\widetilde{\bm X}_i=\bm X_i\bm F$ obtained using the public feature mapping, with $\bm F\in\mathbb R^{d\times h}$ and $\bm F^\top\bm F=\bm I_h$, these attack formulas apply with $(\bm X_i,d)$ replaced by $(\widetilde{\bm X}_i,h)$. The recovered embedded rows $\widehat{\widetilde{\bm X}}_i$ are decoded as $\widehat{\widetilde{\bm X}}_i\bm F^\top$, giving an estimate in the public subspace. For image evaluation, the decoded rows are reshaped to RGB images and clipped to $[0,1]$.

\subsection{Moore--Penrose Estimator}
\label{subsec:supp_mp_attack}

MP removes the orthogonality constraint and solves
\begin{equation}
    \widehat{\bm O}_i
    =\arg\min_{\bm O\in\mathbb R^{d\times d}}
    \left\|\overline{\bm B}_i-\bm A\bm O\right\|_F^2.
    \label{eq:supp_mp_attack_objective}
\end{equation}
Because $\bm A$ has full column rank and $\bm A^\top\bm A=\bm I_d$, this least-squares estimate is unique:
\begin{equation}
    \begin{aligned}
    \widehat{\bm O}_i
    &=\bm A^\dagger\overline{\bm B}_i
    =\bm A^\top\overline{\bm B}_i\\
    &=\bm O_i+v\bm A^\top\bm H_r\bm W_i.
    \end{aligned}
    \label{eq:supp_mp_attack_rotation}
\end{equation}
At $v=0$, the estimate equals $\bm O_i$. 

\subsection{Alignment-Map Calibration Estimator}
\label{subsec:supp_am_attack}

AM calibrates the fitted alignment maps using the colluder's disclosed rotation. At $v=0$, any zero-residual solution of the centered generalized orthogonal Procrustes problem satisfies $\bm A\bm O_j\bm R_j=\bm C$ for every participant. Since $\bm A$ has full column rank, $\bm O_j\bm R_j=\bm Q$ for a common $\bm Q\in\mathcal O(d)$. This motivates
\begin{equation}
    \begin{aligned}
    \widehat{\bm Q}&:=\bm O_c\widehat{\bm R}_c,\\
    \widehat{\bm O}_i
    &:=\widehat{\bm Q}\widehat{\bm R}_i^\top
    =\bm O_c\widehat{\bm R}_c\widehat{\bm R}_i^\top.
    \end{aligned}
    \label{eq:supp_am_attack_rotation}
\end{equation}
AM reconstructs with $\widehat{\bm O}_i^\top$. AM uses the relative alignment between the colluding participant and the protected participant. Applying the same additional orthogonal rotation to all alignment maps does not change the estimate in \eqref{eq:supp_am_attack_rotation}. With noisy anchors, AM uses the maps returned by spectral alignment; these maps need not align the anchors perfectly.

\subsection{Experimental Comparison}
\label{subsec:supp_attack_comparison}

\begin{figure*}[!tbp]
\centering
\includegraphics[width=\textwidth]{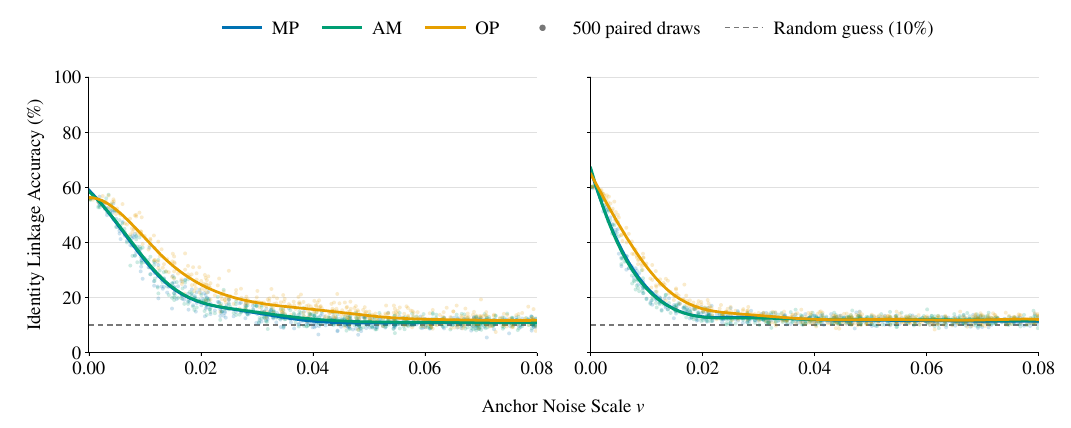}
\caption{Identity linkage under MP, AM, and OP reconstruction attacks on NAA-GDP: the CelebA (left) and VGGFace2 (right) datasets. The dashed line marks $10\%$ random-guessing accuracy.}
\label{fig:supp_attack_comparison}
\end{figure*}

Fig.~\ref{fig:supp_attack_comparison} compares MP, AM, and OP attacks on the CelebA and VGGFace2 datasets using the same $p=10$ data allocation and identity-linkage evaluation as Section V-B of the main manuscript, but a separate uniform noise-draw schedule. We independently sample 500 anchor-noise scales uniformly from $[0,0.08)$ and generate a fresh Gaussian noise realization for each draw. For each draw, MP, AM, and OP are applied to the same uploads and evaluated using the same galleries. Lines are descriptive cubic regression splines with five equally spaced interior knots.

Linkage generally decreases toward random-guessing accuracy as the anchor-noise scale increases. In the separate zero-noise diagnostic, all three attacks attain $55.7\%$ linkage on the CelebA dataset and $60.2\%$ on the VGGFace2 dataset. This diagnostic is excluded from the 500 observations, summary statistics, and fitted curves. 

On the CelebA dataset, mean linkage across the 500 draws is $19.79\%$ for OP, $16.48\%$ for MP, and $16.78\%$ for AM. OP exceeds MP by a mean of $3.31$ percentage points and AM by $3.01$ points. On the VGGFace2 dataset, the corresponding means are $16.66\%$, $15.15\%$, and $15.21\%$. OP exceeds MP by $1.51$ points and AM by $1.45$ points. Thus, OP attains the highest average linkage on both datasets over the prespecified uniform noise range.

OP's advantage is most pronounced in the lower-noise region. On both datasets, OP retains higher identity-linkage accuracy at small, nonzero anchor-noise scales, while linkage under MP and AM decreases more rapidly. As the noise scale increases, the differences narrow and all three attacks approach the $10\%$ random-guessing level. Across the full noise range, OP has strictly higher linkage than both alternatives in 367 of 500 draws on the CelebA dataset and 263 of 500 draws on the VGGFace2 dataset; it ties for the highest linkage in another 3 and 10 draws, respectively. These results support OP as the primary evaluated reconstruction attack in the main experiments, particularly in the lower-noise region where the choice of estimator has the greatest effect on measured leakage. They also show that increasing anchor noise reduces linkage under all three evaluated attacks. The ordering can vary across individual realizations, and the comparison does not establish OP's optimality against other attacks or provide a privacy guarantee.
